\documentclass[twoside]{article}

\usepackage[accepted]{aistats2021}
 
\usepackage[utf8]{inputenc} 
\usepackage[T1]{fontenc}  
\usepackage{hyperref}   
\usepackage{url}      
\usepackage{booktabs}   
\usepackage{amsfonts}  
\usepackage{nicefrac}    
\usepackage{microtype}    

\usepackage{graphicx}
\usepackage{wrapfig}
\usepackage{soul}
\usepackage{subcaption}
\usepackage{amssymb}
\usepackage{amsmath}
\usepackage{amsthm}
\usepackage{algorithm}
\usepackage{algorithmic}
\usepackage{multirow}
\usepackage{chngpage}

\usepackage{amsmath,amsfonts,bm}

\def\eqref#1{equation~\ref{#1}}

\def\1{\bm{1}}

\def\vx{{\bm{x}}}
\def\vy{{\bm{y}}}

\def\mH{{\bm{H}}}
\def\mI{{\bm{I}}}
\def\mJ{{\bm{J}}}

\def\mX{{\bm{X}}}

\DeclareMathAlphabet{\mathsfit}{\encodingdefault}{\sfdefault}{m}{sl}
\SetMathAlphabet{\mathsfit}{bold}{\encodingdefault}{\sfdefault}{bx}{n}

\newcommand{\KL}{D_{\mathrm{KL}}}

\usepackage{xcolor}
\definecolor{linkcolor}{RGB}{0,128,255}
\usepackage{hyperref}
\hypersetup{
    colorlinks=true,
    allcolors=linkcolor
}

\definecolor{dark-blue}{rgb}{0.15,0.15,0.4}

\hypersetup{
  colorlinks=true,
  linkcolor=dark-blue,
  citecolor=dark-blue,
  urlcolor=black,
}

\usepackage[round]{natbib}

\usepackage{xcolor}

\begin{document}

\runningauthor{Maddox, Tang, Moreno, Wilson, Damianou}

\twocolumn[

\aistatstitle{Fast Adaptation with Linearized Neural Networks}

\aistatsauthor{ 
	Wesley J. Maddox$^{*}$ \\
	New York University \\
	\texttt{wjm363@nyu.edu}
	
	\\\And 
	Shuai Tang$^{*}$ \\
	UC San Diego \\
	\texttt{shuaitang93@ucsd.edu}

	\\\And
	Pablo Garcia Moreno \\
	Amazon, Cambridge \\
	\texttt{morepabl@amazon.com}
	
	\\\AND
	Andrew Gordon Wilson \\
	New York University \\
	\texttt{andrewgw@cims.nyu.edu}
	
	\\\And 
	Andreas Damianou\\
	Amazon, Cambridge \\
	\texttt{damianou@amazon.com}
	
	 }
	\begin{center} 
	\footnotesize$^*$Work partly completed during internship at Amazon.
	\end{center}
	\noindent \\
]

\begin{abstract}
	The inductive biases of trained neural networks are difficult to understand and, consequently, to adapt to new settings. We study the inductive biases of linearizations of neural networks, which we show to be surprisingly good summaries of the full network functions. Inspired by this finding, we propose a technique for embedding these inductive biases into Gaussian processes through a kernel designed from the Jacobian of the network. In this setting, domain adaptation takes the form of interpretable posterior inference, with accompanying uncertainty estimation. This inference is analytic and free of local optima issues found in standard techniques such as fine-tuning neural network weights to a new task. We develop significant computational speed-ups based on matrix multiplies, including a novel implementation for scalable Fisher vector products. Our experiments on both image classification and regression demonstrate the promise and convenience of this framework for transfer learning, compared to neural network fine-tuning. Code is available at \url{https://github.com/amzn/xfer/tree/master/finite_ntk}.
\end{abstract}

\section{INTRODUCTION}

Deep neural networks (DNNs) trained on a source task can be used for predicting in a new (target) task through a process which we interchangeably refer to as transfer learning or domain adaptation \citep{tricks_of_the_trade,bengio_deep_2012,yosinski_how_2014,sharif_razavian_cnn_2014}. 
One family of approaches to solve this problem adapts the parameters of the full source task network, through gradient descent or more elaborate methods such as meta-learning \citep{finn_model-agnostic_2017}. 
However, adaptation of the network is computationally demanding and prone to getting trapped in local minima. 
Moreover, fine-tuning the full network in excess can lose a useful representation that was learned from the source domain. 

A compelling alternative is to fine-tune only the last layer of the source network, which leads to a log-convex problem that avoids local optima issues and, at the same time, provides a more computationally affordable and less data demanding solution. 
However, the resulting solution can lead to a poor fit when dealing with complex transfer learning tasks due to the lack of flexibility. 
Moreover, it is not clear how many of the last layers one would need to fine-tune and what are the overall inductive biases that are transferred to the target task in each case after these layers' parameters have been moved from their original states.

In this paper we aim to address the limitations of fine-tuning with minimal computational overhead, while keeping the performance competitive with costly and involved solutions based on adapting the full network. We propose to structure the transfer learning problem in the following simpler way: firstly, we linearize the DNN with a first order Taylor expansion, giving rise to a linear model whose inductive biases we study here empirically. Secondly, we embed these inductive biases into a probabilistic lightweight framework which takes the form of a Bayesian linear model with the DNN Jacobian matrix $\mJ$ as the features or, equivalently, a Gaussian process (GP) with $\mJ^\top \mJ$ as the kernel. The kernel matrix is the finite width counterpart of the neural tangent kernel (NTK) \citep{jacot_neural_2018,lee_wide_2019} but, crucially, our Taylor expansion is performed around a \emph{trained} model, rather than a randomly initialized one. 
We then achieve fast adaptation by computing the Jacobian matrix on the target task and invoking the GP predictive equations. 

\begin{figure}[t]
	\centering
	\includegraphics[angle=270,width=\linewidth,clip,clip,trim=2.5cm 0cm 2cm 3cm]{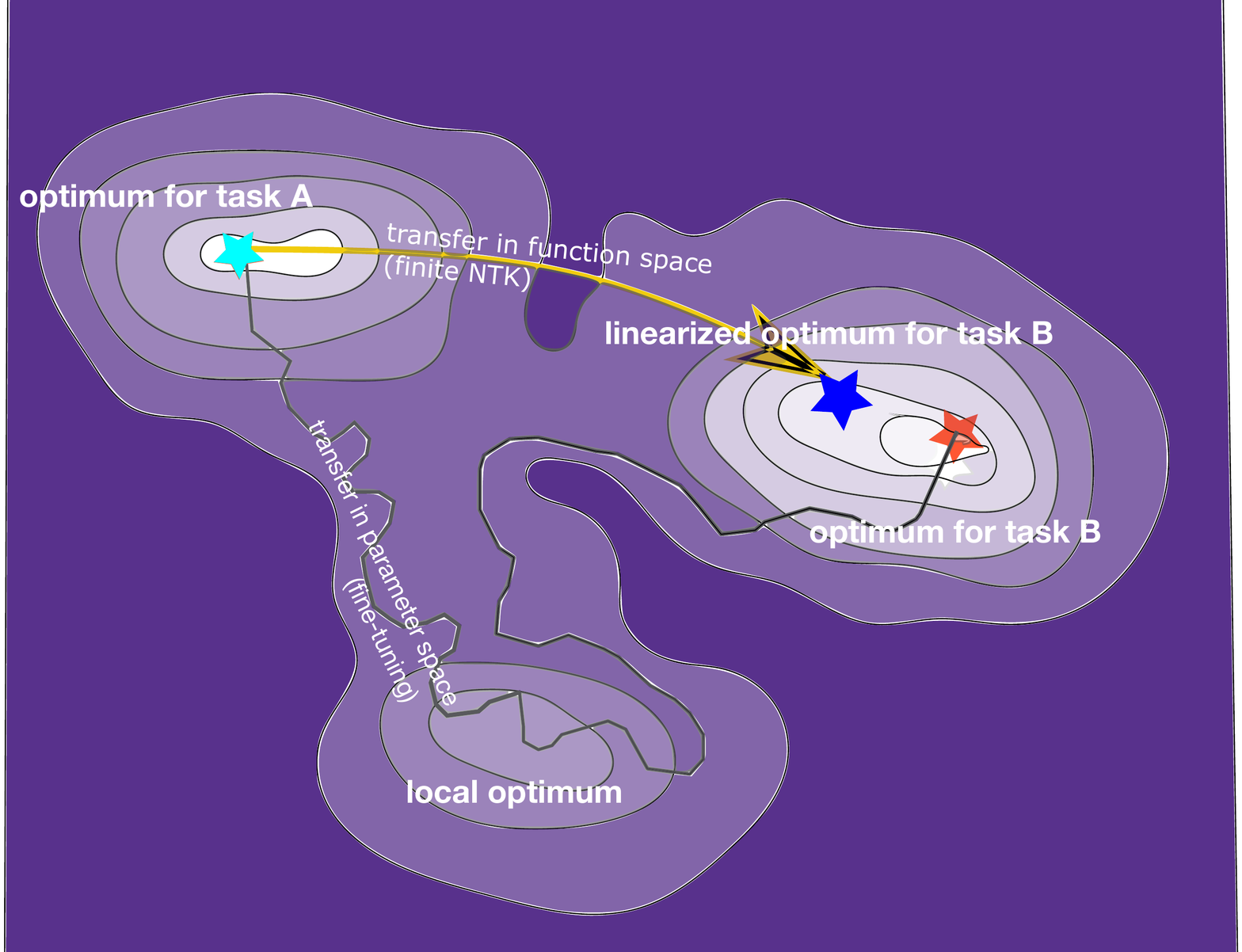}
	\caption{Schematic of our approach to probabilistic transfer learning. We show the loss surfaces for the model as we transfer from the source task (light blue star) to the target task (red/blue stars). Linearizing the network allows for transferring in \emph{function} space which gives a direct mapping to an optimum, while transferring in parameter space (e.g. fine-tuning) can get stuck in local optima.}
	\label{fig:transfer_optima}
\end{figure}

We therefore cast domain adaptation as posterior inference in function space. This solution is analytic (and closed-form, for regression), hence by-passing local optima issues. It is also interpretable: firstly, all of our prior assumptions are encoded in the kernel of the GP, whose inductive biases we study; secondly, our probabilistic model gives a calibrated estimate of the uncertainty of the transfer task and closed form predictive distributions for regression. 
Figure \ref{fig:transfer_optima} sums up our approach to transfer learning. 
Given a trained model at a global optimum (purple star), we transfer in function space using the linearized model to another optimum (red star) avoiding local optima issues incurred by transferring in parameter space using fine-tuning. 
 
To make our approach competitive with respect to non-probabilistic domain adaptation methods we need to tackle one remaining challenge: scalable inference. To this end, we consider a black box inference framework which uses implicit Jacobian vector products and, hence, avoids forming the kernel explicitly. This is complemented by a novel implementation of a directional derivative based on Fisher vector products to enable fast black-box conjugate gradients optimization and test-time caching for fast predictive variances \citep{gardner_gpytorch:_2018,pleiss_constant-time_2018}. 
Overall, we are thus able to combine the benefits of probabilistic and neural network models without sacrificing scalability. 

Historically, linearizing a sophisticated non-linear trained model dates back at least several decades in machine learning, including Laplace approximations \citep{mackay_bayesian_1992}, and perhaps most notably the Fisher kernel \citep{jaakkola_exploiting_1998}, and related approaches involving Gaussian processes \citep{seeger2002covariance}.  
	Interest has recently reignited in these approaches. Beyond the original version of this paper \citep{maddox_linearizing_2019}, \citet{khan_approximate_2019} also used linearized networks, but for tuning hyper-parameters of neural network architectures.
	Specifically, they used weight parameters derived from variational interpretations of stochastic gradient descent.
	In other work, \citet{mu_gradients_2019} use some of the gradients of unsupervised deep models to transfer to labelled data. Both approaches share some aspects of our construction, using the neural tangent kernel at finite width with trained networks, rather than networks at initialization \citep{jacot_neural_2018}.

We consider applications in both regression and classification. Our key contributions are as follows: 
\begin{itemize}
	\item We study empirically the inductive biases of the linearized DNN model, demonstrating that the linearized model still maintains the strong capabilities of the full neural network for transfer learning. 
	\item We include neural networks in a probabilistic framework by considering the aforementioned linearized DNN model. 
	We develop techniques for fast inference, including a novel implementation for scalable Fisher vector products.
	\item We apply this probabilistic framework to domain adaptation, casting transfer learning as posterior inference in function space. We demonstrate that this produces strong performance compared to fine-tuning. 
\end{itemize}

The rest of the paper is organized as follows: in Section \ref{sec:rel_work} we summarize related work using the Jacobian matrix of neural networks, in Section \ref{sec:GP_NTK} we describe how we efficiently use the Jacobian matrix in either a Bayesian (generalized) linear model or a degenerate Gaussian process. In Section \ref{sec:ind_biases}, we test the inductive biases of these Jacobian kernels. Finally, in Section \ref{sec:exps}, we test their domain adaptation abilities, demonstrating that they are a strong principled baseline for fast adaptation.

\section{RELATED WORK}\label{sec:rel_work}

Using the gradients of a model (such as a DNN) as features for classification problems has a long history. The Fisher kernel used both the Jacobian and inverse Fisher matrices of a generative unsupervised model as the kernel for support vector machines (SVMs) \cite{jaakkola_exploiting_1998}.
More recently, \citet{zhai_adversarial_2019} constructed Fisher vectors from GANs and used them to produce highly accurate linear models on CIFAR10.
\citet{zinkevich2017holographic} demonstrated that linear models can be constructed from the Jacobians of DNNs to match the predictions of the full DNN on the training set.
\citet{tosi_metrics_2014} and \citet{tosi_visualization_2014} interpreted the Jacobian matrices for different tasks as creating a probabilistic metric tensor whose expectation is the average Fisher information across several tasks. Finally, we note that model linearization using the Jacobian can be linked to 
Laplace approximation \citep{mackay_information_2003}.

Empirical evidence has shown that features across neural networks are transferable \citep{bengio_deep_2012}.
In particular, \citet{yosinski_how_2014} and \citet{li_convergent_2015} found that features learned by neural networks were transferable across both tasks and architectures. 
These empirical results motivate our work as the Jacobian is a transformation of the features of the network.
More recently, (diagonal) Fisher matrices for classifiers have been shown to be informative for estimating similarity across tasks \citep{achille_task2vec:_2019}.
This paper is a longer version of our original work \citep{maddox_linearizing_2019}, which linearizes trained neural networks for 
transfer learning.
\citet{mu_gradients_2019} also consider transfer learning, arguing that the Jacobian is useful for representation learning because it is a local linearization of fine-tuning.
They use the Jacobian of several layers of the neural network, rather than the whole network, to transfer deep unsupervised models to become supervised ones.
We compare to their approach in Section \ref{sec:exps}.

\citet{jacot_neural_2018} showed that infinitely wide DNNs behaved as their associated Taylor expansion around \textit{initialization}, terming the resultng kernel in the infinite limit the neural tangent kernel (NTK).
\citet{lee_wide_2019} extended their analysis showing that the evolution of finitely wide DNNs over the course of training is similar to that of linear models. 
\citet{lee_wide_2019} and \citet{arora_exact_2019} gave implementations of infinitely wide DNNs, demonstrating strong performance on CIFAR-10.
Due to the analytic nature of these computations, the DNNs studied are restricted: allowing no batch normalization and a slim choice of layers, and use a different scaling than standard DNNs.

\citet{khan_approximate_2019} use Jacobian features and a variational distribution in weight space derived from a linear model to tune neural network hyperparameters. They also exploit the GP kernel trick to turn their linear model into a GP, but due to computational concerns only use the diagonal of the Jacobian. After the initial publication of our work, \citet{pan_continual_2021} proposed a similar function space approach but for continual learning that uses the Jacobian matrix and a Laplace approximation. Similarly, \citet{immer_improving_2021} have recently proposed an approach to exploit generalized linear models and the linearization to improve predictions.

\section{LINEARIZED NEURAL NETWORK MODEL}
\label{sec:GP_NTK}

We first relate using the Jacobian matrix as features to degenerate Gaussian processes.
Given an arbitrary neural network, $f,$ with $p$ parameters $\theta$ and $n$ inputs $x = \{x_i\}_{i=1}^n$ with outputs in dimension $o,$ we may Taylor expand $f$ around $\theta$ in the following manner:
\begin{align}
f(x; \theta') \approx f(x; \theta) + \mJ_\theta(x)^\top (\theta - \theta'),
\label{eq:taylor}
\end{align}
where $\mJ_\theta(x)$ is the 
	$p \times on$
	Jacobian matrix of the model.\footnote{We will drop the dependency on $x$ and squeeze $o$ outputs.}
We will term Eq. \ref{eq:taylor} the \emph{linearized NN} model and will consider additionally the 
first-order approximation 
as their own model, $g(x; \theta') \approx \mJ_\theta(x)^\top (\theta - \theta').$ 
Interestingly, this model can be seen as a version of the neural tangent kernel \citep{jacot_neural_2018} but at finite width, so we refer to it as the finite NTK. 
For regression, the two models (finite NTK and linearized NN) only differ in the choice of the mean function, which is somewhat unimportant in comparison to the covariance, so we will focus our experiments on the finite NTK.

{As our primary goal is to account for functional uncertainty in new tasks, we will consider the linearized model in a Bayesian setting, assigning a prior to the linearized parameters, $\theta'.$}
We assume that $\theta' \sim \mathcal{N}(0, \mI_p)$ for simplicity as in \citet{jacot_neural_2018}; however, more highly structured Gaussian priors could be used.
Under the Gaussian prior assumption, the linearized model and the gradient model are degenerate Gaussian processes (so named because in function space, the resulting kernel is degenerate --- i.e. it has a finite number of features) \citep{rasmussen_gaussian_2008}.
Both models have the same kernel, $k(x_i,x_j) = \mJ_\theta(x_i)^\top \mJ_\theta(x_j),$ but different mean functions.
In contrast to \citeauthor{jacot_neural_2018}, we take the Taylor expansion around a trained neural network rather than the network at initialization.

\subsection{Function Space Updates for Regression}\label{sec:basis_fn}

Following \citet{rasmussen_gaussian_2008}, the predictive posterior over the function $f^*$ on new inputs $x^*$ is:
\begin{align}
	f^* | x^*, \mathcal{D} \sim \mathcal{N} &\left( \mJ_\theta^{*\top} (\mJ_\theta \mJ_\theta^\top + \sigma^2 \mI_p)^{-1}  \mJ_\theta \vy, \right. \nonumber\\
	& \left. \sigma^2 \mJ_\theta^{*\top} (\mJ_\theta \mJ_\theta^\top + \sigma^2 \mI_p)^{-1} \mJ_\theta^{*}  \right)
	\label{eq:parameter_space_inference}
\end{align}	
where $\mJ_\theta^*$ is the Jacobian matrix on the new data points.\footnote{Dropping the dependency on $x^*$ for a superscript and using $\vy$ to include both the response and the mean terms $\mu_\theta(x) = \mJ_\theta^\top \theta + f_\theta(x)$.}
We can now clearly see that inference (and predictive variance computation) only involves inverting a $p \times p$ matrix, e.g. \textit{solving the linear system} $(\mJ_\theta \mJ_\theta^\top + \sigma^2 \mI_p) x = b,$ with the Gram matrix, $\mJ_\theta \mJ_\theta^\top.$
Naively, this operation requires $\mathcal{O}(p^3)$ time as the linear system is solved in parameter space (the weight space view).
Fortunately, we can flip the computations in the dual function space by interpreting $\mJ_\theta$ as producing a degenerate Gaussian process with kernel matrix $\mJ_\theta^\top \mJ_\theta.$
Using Woodbury's matrix identity, the posterior predictive distribution can be re-written as:
\begin{align}
	f^*& | x^*, \mathcal{D} \sim \mathcal{N}\left(
	\mJ_\theta^{*T} \mJ_\theta(\mJ_\theta^\top \mJ_\theta + \sigma^2 \mI_n)^{-1} \vy, \right. \nonumber\\ 
	&\left. \sigma^2 \mJ_\theta^{*T} (\mI_p - \mJ_\theta (\mJ_\theta^\top\mJ_\theta + \sigma^2 \mI_n)^{-1} \mJ_\theta^\top) \mJ_\theta^* \right) 
	\label{eq:gp_pred_dist}
\end{align}
Again, inference only requires solving the linear system $(\mJ_\theta^\top \mJ_\theta + \sigma^2 \mI_n) x = b$, naively requiring $\mathcal{O}(n^3)$ 
time as the linear system is solved in function space (the function space view). Popular neural networks are generally over-parameterized, containing many more parameters than training samples; therefore,
the function-space view will typically be faster. In the following section, we will consider computational considerations.

\begin{figure*}[t!]
	\begin{subfigure}{0.6\textwidth}
		\includegraphics[width=\linewidth]{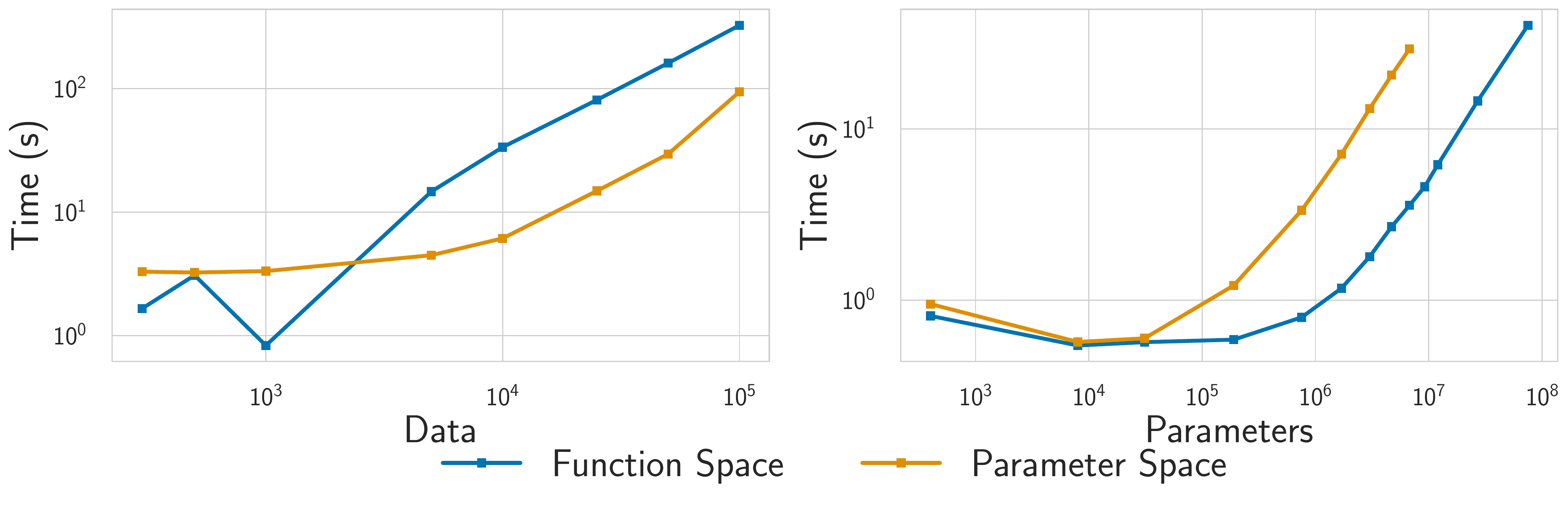}
	\end{subfigure}
	\begin{subfigure}{0.3\textwidth}
		\includegraphics[width=\linewidth]{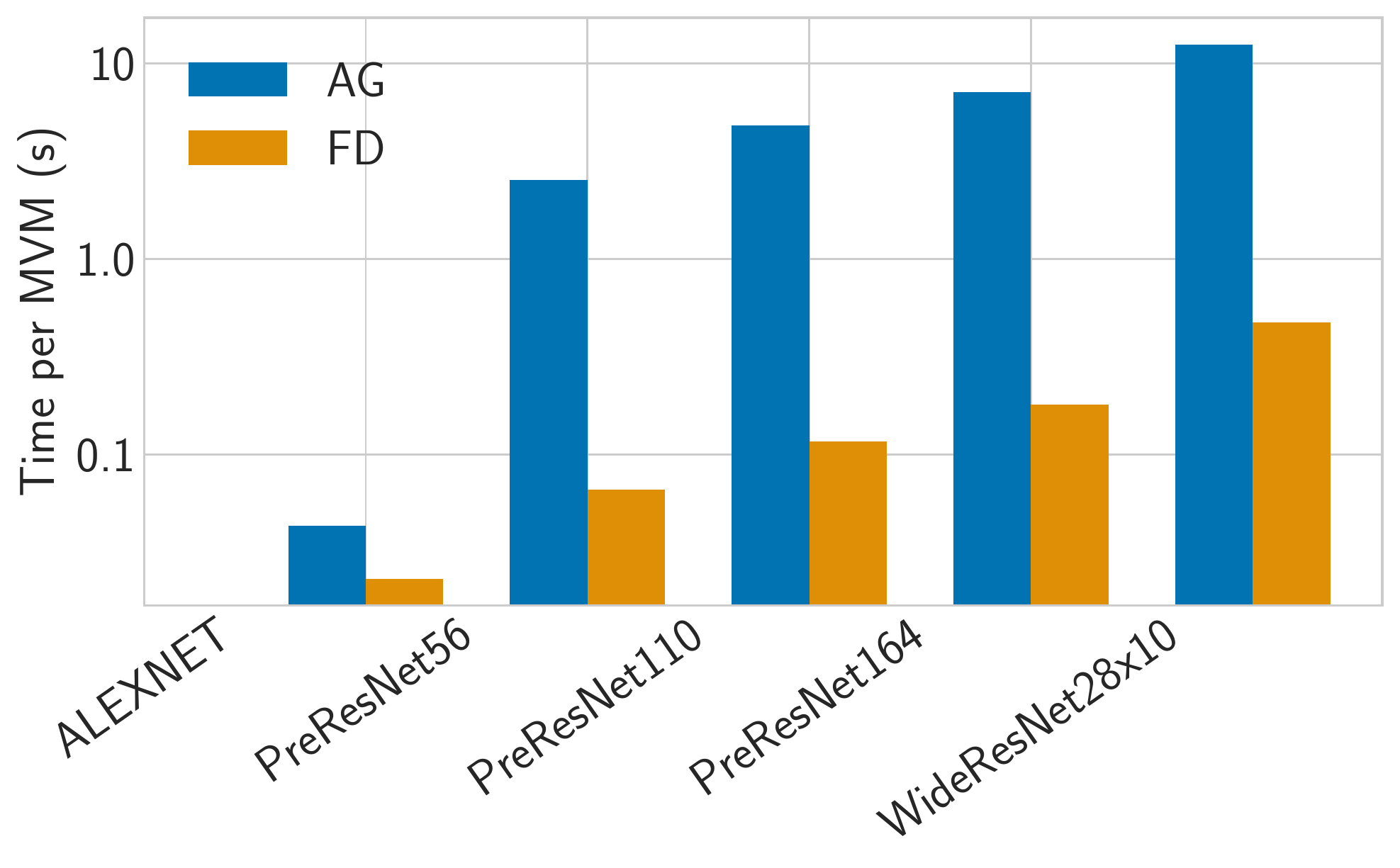}
	\end{subfigure}
	\centering
	
	\caption{\textbf{Left:} Nearly linear scaling time of log probability calculation for an $\approx 800,000$ parameter MLP as a function of $n.$  \textbf{Center:} Nearly linear scaling time of log probability calculation for $1,000$ data points as a function of model size. In both situations, both the function-space approach and the parameter space approach scale particularly well, reaching $100,000$ data points (and nearly 100 million parameters) unlike standard GP training procedures which are $\mathcal{O}(n^3).$ {
			\textbf{Right:} {Speedup of finite differences (FD) Fisher-vector products against autograd (AG) for AlexNet \citep{krizhevsky_imagenet_2012}, PreResNets \citep{he_deep_2016} of varying depth, and WideResNet \citep{huang_densely_2017} on CIFAR-10. Our method typically achieves a $30\times$ speedup over the standard implementation, demonstrating the practical utility of our approach to working with the Fisher matrix.}
		}
	}
	\label{fig:ntk_speed}
\end{figure*}

\paragraph{Extension to Non Gaussian Likelihoods:}
For non-Gaussian likelihoods, a degenerate Gaussian process on the latent functions is again produced, equivalent now to a Bayesian generalized linear model (GLM). 
Specifically, for multi-class classification the GLM contains the Jacobian within a categorical likelihood: 
\begin{align}
p_{\text{lin}}(y_i | x, \theta) = \text{Cat.}\left(y_i | \frac{\exp\{\mJ_\theta^\top \theta'\}}{\sum_{i=1}^C \exp\{\mJ_\theta^\top \theta'\}}\right),
\label{eq:cat_lin}
\end{align}
with appropriate reshaping to account for the $on$ outputs of the model.
Unfortunately, the posterior over $\theta'$ cannot be computed in closed-form.  
We consider both Laplace approximations and stochastic variational inference \citep{hoffman_stochastic_2013}.
See Appendix \ref{app:approxinf} for details.

\subsection{Computational Speed-Ups}\label{sec:computational}
To consider large DNNs within our approach, we need to tackle two scalability issues: firstly the Jacobian matrix is not closed form for most NN models, and secondly, GP inference scales poorly. 
We resolve both issues simultaneously by using implicit Jacobian vector and vector Jacobian products as implemented in standard automatic differentiation software like Pytorch \citep{paszke2019pytorch}.
Implicit Jacobian vector products never form the full Jacobian matrix and use three backwards calls, one to compute $\mJ_\theta^\top v$ and two for $\mJ_\theta v,$ rather than $n.$
They are also extremely compatible with conjugate gradient (CG) approaches for GP inference (e.g. GPyTorch), as CG only requires matrix vector multiplications, resolving the poor scaling of GP inference.
CG-enabled GP inference reduces the inference complexity from $\mathcal{O}(n^3)$ to $\mathcal{O}(n^2),$ while keeping the memory required constant by not forming dense matrices unnecessarily \citep{gardner_gpytorch:_2018}.
More implementation details are in Appendix \ref{app:fast_gps}.

To demonstrate the efficiency of combining CG-enabled GP inference with implicit Jacobian vector products, we performed experiments with up to $100,000$ data points using five layer perceptrons with about $800,000$ parameters using only a single GPU. We show the timing in Figure \ref{fig:ntk_speed},  finding that predictive means and variances can be computed in under five minutes on even the largest models and datasets.
We used similar architectures in Section \ref{sec:exps}.

For further computational improvements, we can perform inference in parameter space by leveraging the Jacobian's relationship to the Fisher information matrix, which is defined as.
	$
	\mathbb{F}(\theta) := \mathbb{E}_{p(x,y|\theta)} (\nabla_\theta \log p(y | x, \theta) \nabla_\theta \log p(y | x, \theta)^\top ).
	$
For Gaussian likelihoods, the Fisher information matrix is proportional to the outer product of the Jacobian matrix with itself, e.g. $\mJ \mJ^\top \propto \mathbb{F}(\theta),$ across a dataset. 

To exploit the relationhip, we derived a novel finite differences Fisher vector product (FVP) via the derivative of the Kullback-Leiber (KL) that gives matrix vector products with only one backwards call. 
The second order Taylor expansion of the KL divergence between two distributions, $p(y | \theta)$ and $p(y|\theta'),$ with parameters $\theta$ and $\theta'$ is given by:
\begin{align*}
\KL(p(y\mid\theta)||p(y\mid\theta')) = &\frac{1}{2}(\theta - \theta')^\top\mathbb{F}(\theta)(\theta - \theta')  + \\
			&\mathcal{O}{(\theta - \theta')^{3}} .
\end{align*}
Evaluating the derivative at $\theta' = \theta + \epsilon v$ gives:
\begin{align}
\nabla_\theta \KL(p(y\mid\theta)||p(y\mid\theta')) |_{\theta' = \theta + \epsilon v} =&\epsilon \mathbb{F}(\theta) v + \nonumber \\
&\mathcal{O}(\epsilon^2 ||v||) 
\label{eq:kl_deriv_fvp}.
\end{align}
Therefore, to compute Fisher vector products, we merely need to compute a second forwards pass with $\theta',$ compute the KL divergence in likelihood space between the model with parameters $\theta$ and the model with parameters $\theta',$ and then backpropagate with respect to $\theta'.$
The resulting gradient, up to error, is proportional to $\mathbb{F}(\theta) v.$
To use these FVPs to increase the speed of our GP implementation, we use CG to solve systems of equations in parameter space (e.g. Eq. \ref{eq:parameter_space_inference}) replacing matrix vector multiplications of the form $\mJ_\theta \mJ_\theta^\top v$ with $a \mathbb{F}(\theta) v,$ where $a$ is the appropriate scaling constant. 
\begin{figure*}[!t]
	\begin{subfigure}{0.6\textwidth}
		\includegraphics[width=\linewidth]{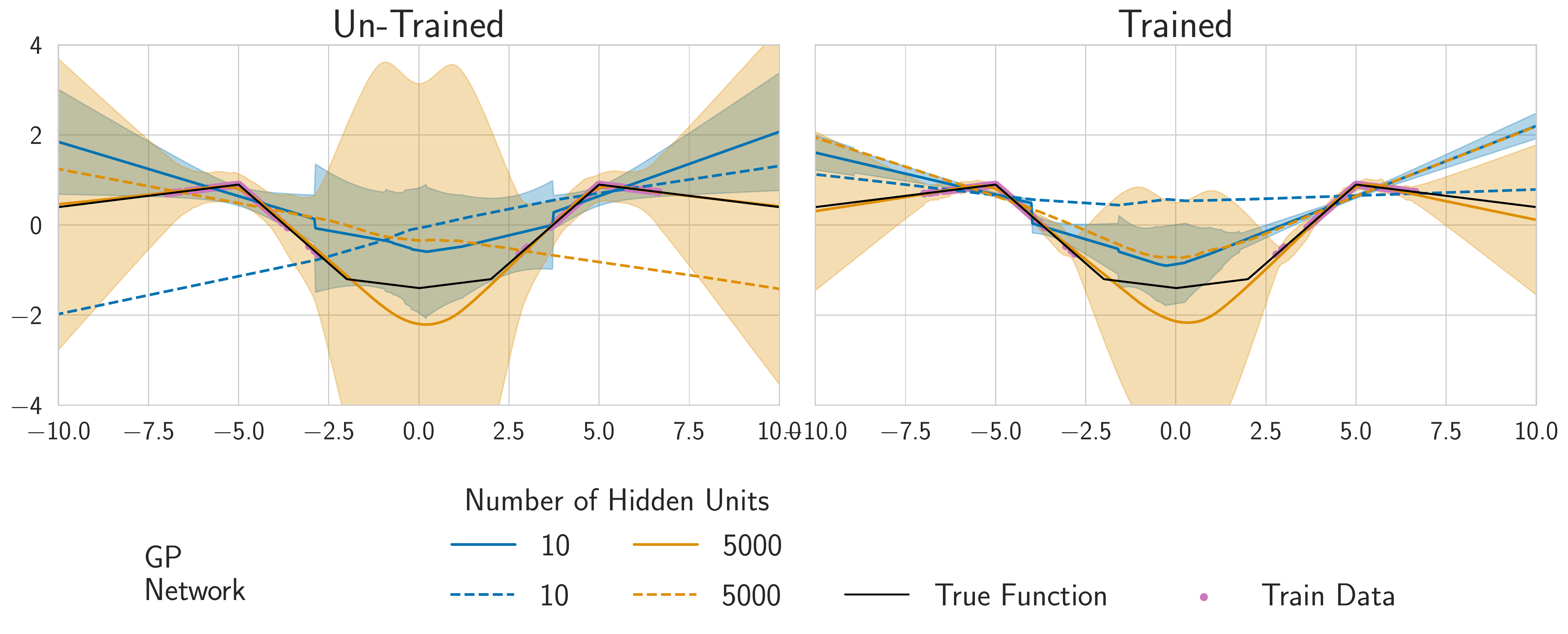}
		\caption{Trained and untrained ReLU networks with one hidden layer.}
		\label{fig:increasing_width}
	\end{subfigure}
	\hfill
	\begin{subfigure}{0.38\textwidth}
		\includegraphics[width=\linewidth]{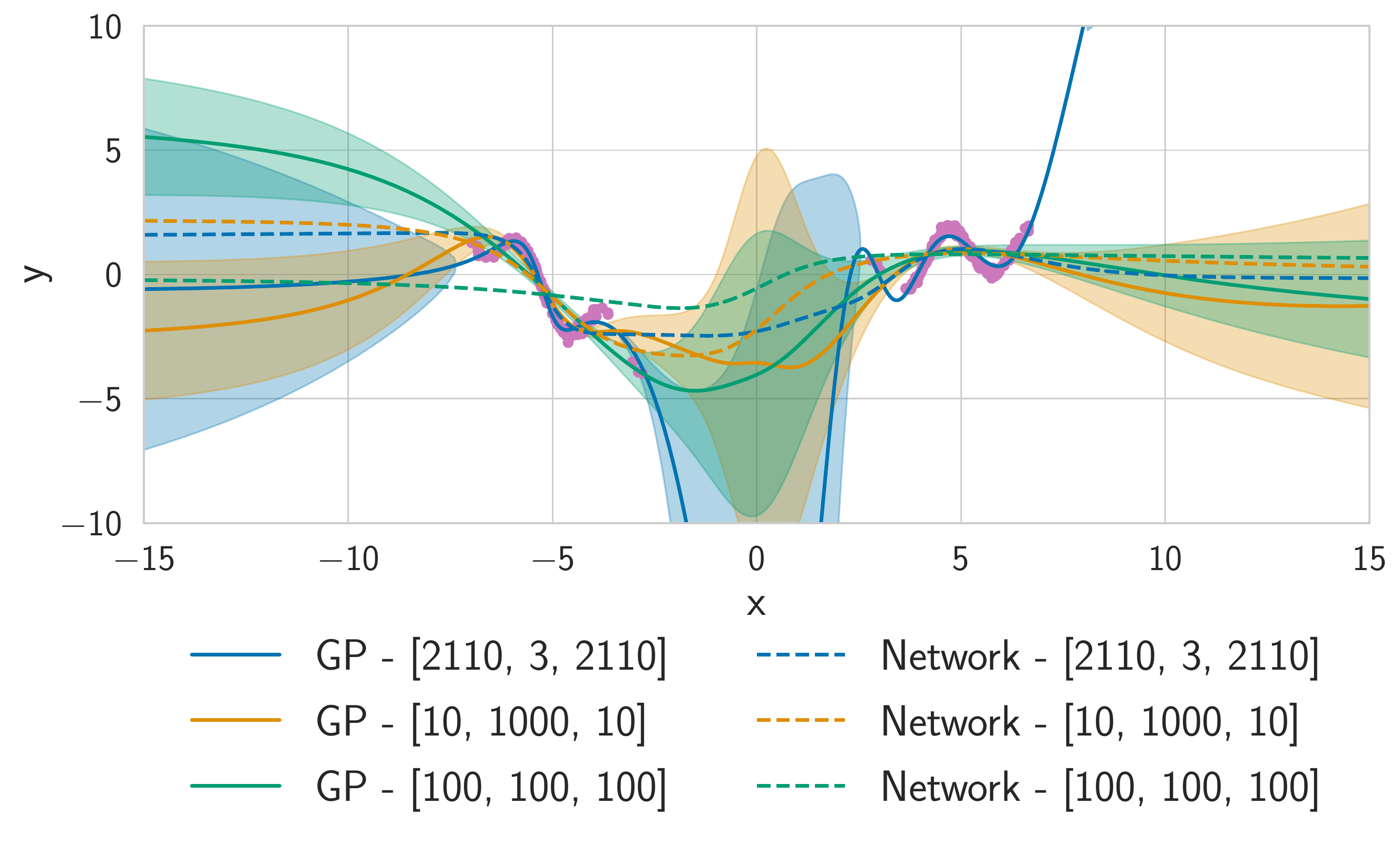}
		\caption{Trained tanh networks of different widths.}
		\label{fig:different_architectures}	
	\end{subfigure}
	\caption{Posterior means and confidence regions, $p(f^* | y),$ for MLPs of varying width and depth. Solid lines are the predictive means from the linearized neural network, with shading corresponding posterior confidence region; dashed lines are predictions from the full neural network. The observed data is shown as pink dots. \textbf{Left:} Linearization of un-trained single-layer ReLU networks of varying width. \textbf{Center:} Linearization for trained networks. Observable in both plots is the increase in the GP predictive variances as a function of width due to using standard initializations, while the trained networks have less predictive variance, especially around the training data (due to being fit on the data). \textbf{Right:} Three layer tanh architectures with different widths but similar numbers of parameters trained on a sinusoidal regression problem. Not only are the network's predictions qualitatively different, but their linearizations also are significantly different outside of the training data.}
\end{figure*}
In Figure \ref{fig:ntk_speed}, we show the efficiency of finite NTK inference using these finite differences (FD) FVPs in comparison to kernel space inference.
We also show the speedup against an exact FVP computed using autograd (AG), where the FD version is about $30\times$ faster across a variety of modern architectures. 
Accuracy is not affected as relative error typically is on the order of $0.001$. Exact implementation details are in Appendix \ref{app:fast_fvp}.

\subsection{Fast Adaptation Modelling} \label{sec:probmodel}
\begin{figure}[h!]
	\begin{minipage}{\linewidth}
\begin{algorithm}[H]
	\caption{Domain Adaptation Procedure}
	\label{alg:fast_adapt}
	\begin{algorithmic}
		\STATE {\bfseries Input:} Data $(\mX_1, \vy_1)$, Initial parameters $\theta_0$
		\STATE Compute $\theta_{MLE}$ with data $(\mX_1, \vy_1).$
		\STATE Transfer inductive biases to GP: \\
		$\ \ f \sim \mathcal{GP}(0,k=\mathbf{J}(\mathbf{x}_1; \theta_{MLE})^\top\mathbf{J}(\mathbf{x}_1; \theta_{MLE}))$
		\STATE Compute Jacobian for task $t$: $\mathbf{J}_t = \mathbf{J}(\mathbf{x}_t; \theta_{MLE})$
		\STATE Adapt $f$ to domain $t$ using Eq. \ref{eq:gp_pred_dist} and $\mathbf{J}_t$: \\
		$\ \ p(f^*_t | \mathcal{D}_t) = \int p(f^* | \theta'_t) p(\theta'_t | \mathcal{D}_t) d\theta'$ 
	\end{algorithmic}
\end{algorithm}
\end{minipage}
\end{figure}

There is a long history of multi-task Gaussian process models, see \citet{alvarez2012kernels} for further discussion.
We adopt the simplest multi-task model, which considers
the parameters of the trained neural network to be shared across all tasks, and recomputing the Jacobian for each new task (equivalent to sharing the kernel hyper-parameters across tasks).

The generative process
can be described as follows:
first, the dataset is drawn from a dataset distribution $\mathcal{D}_t = (x_t, y_t) \sim p(\mathcal{D})$, so that, for each individual task and dataset, $\mathcal{D}_t$, we have the likelihood $p(\vy_t | x_t) = p(\vy_t | f_{\theta_t}(x_t))$.
In our setting, we have the linearized neural network, $f_t \approx \mJ_{\theta}(x_t) \theta'_t + \mu(x_t),$ where $\mJ_\theta(x) = (\nabla_\theta f(x))^\top \in \mathbb{R}^{p \times on},$ and $\mu(x,t)$ is the GP mean function as described previously,
and $\theta'_t$ are the parameters of the (Bayesian) linearized model.
The kernel for each task is given by $\mJ_\theta(x_t)^\top \mJ_\theta(x_t')$ as in Section \ref{sec:basis_fn}.
Note that the Jacobian computation depends only on the parameters $\theta$ of the pre-trained network. We can then represent the functions $f_t$ (for other tasks) in a way that does not involve a new non-convex optimization. Thus, our procedure only requires a pretrained neural network on an initial task. When a new task is presented, the Jacobian matrix of data samples in the given task w.r.t. the pretrained parameters is used to derive the predictive distribution of the GP, which only requires solving a linear system. 
We present the overall algorihtm in Algorithm \ref{alg:fast_adapt}. In Appendix \ref{app:fast_adaptation} we further describe the domain adaptation model.

\section{INDUCTIVE BIASES OF LINEARIZED NEURAL NETWORKS}\label{sec:ind_biases}
We first investigate the inductive biases of linearized neural networks on both regression and classification on CIFAR10, while showing that the Jacobians of similar tasks are related.
Overall, we demonstrate i) that linearized neural networks retain the inductive biases of their nonlinear counterparts and ii) that the Jacobian matrix is a useful inductive bias for transfer learning. 
\begin{figure*}[!t]
	\centering
	\includegraphics[width=0.9\textwidth,clip,clip,trim=0cm 7cm 0cm 7cm]{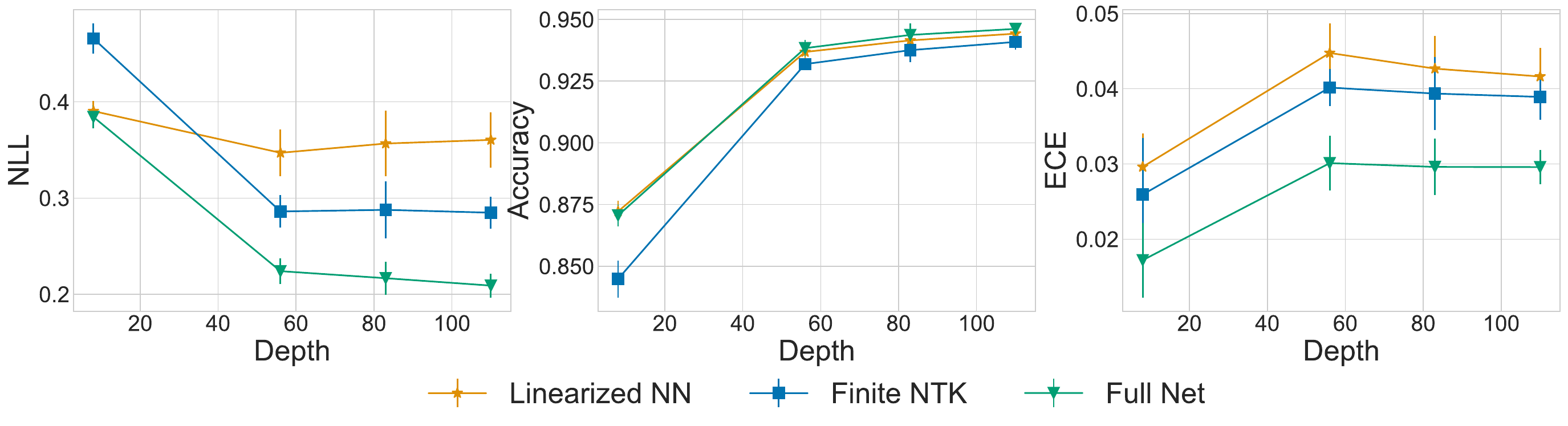}
	\caption{Test negative log-likelihood (NLL), accuracy, and expected calibration error (ECE) of PreResNets varying depth and their linearized counterparts as a function of depth on CIFAR-10, averaged over 10 seeds. For the linearization, we use a VI approach. Both the linearized NN (orange) and the finite NTK (blue) perform well in terms of accuracy, nearly comparable to the full network (green). However, they perform slightly worse relatively in terms of NLL and ECE, implying that they are more over-confident about their predictions. }
	\label{fig:prn_c10}
\end{figure*}
\paragraph{Qualitative Regression Experiments:}\label{sec:ind_regression}
We begin with demonstrating qualitatively that linearized neural networks retain good inductive biases.
See Appendix \ref{app:regression_exp_details} for training and dataset details.
First, in Figure \ref{fig:increasing_width}, we show the fit of the ReLU network trained on a synthetic dataset, alongside the posterior predictive mean and variance of the resulting Gaussian process.
We do so for both trained and un-trained networks. Pleasingly, in both cases, having enough hidden units (in this case, $5000$) gives reasonable predictions.

Second, in Figure \ref{fig:different_architectures}, we show the effect of different architectures on a different synthetic dataset. 
We can again qualitatively see that architectures that have good fits to the data (e.g. not over-fit) tend to also have reasonable predictive means and variances (e.g. the predictive mean fits the training data well and the predictive variance increases away from the data).
We choose three separate architectures that all have about $21,000$ parameters, but differing numbers of hidden units at each layer to test the hypothesis that the performance of the linearized NN is tied to the over-parameterization effect rather than the architecture itself.
Given that performance vastly differs even in this simple setting, we attribute the performance to the architecture itself, rather than over-parameterization.
\vspace{-0.25cm}
\paragraph{Linearized PreResNets on CIFAR-10:}
We next tested the accuracy, test loglikelihood (NLL), and expected calibration error (ECE) \citep{guo_calibration_2017} for PreResNets of varying depth on CIFAR10, displaying these results in Figure \ref{fig:prn_c10}.
Here, using the Jacobian matrices as features for classification leads to highly competitive performance for Bayesian generalized linear models --- about 92\% with our VI approximation for the standard $56$ layer version. By comparison the top known kernel method (infinite un-trained neural network, but with a different architecture) on CIFAR-10 seems to be the result of \citet{li_enhanced_2019}, which is about 89\%. 
Further results with MAP inference and Laplace approximations are in Appendix \ref{app:transfer_exps}.
Intriguingly, we also find that the linearized networks here are more-overconfident (higher ECE) than the full model.
\begin{figure}[t!]
	\centering
	\includegraphics[width=0.95\linewidth]{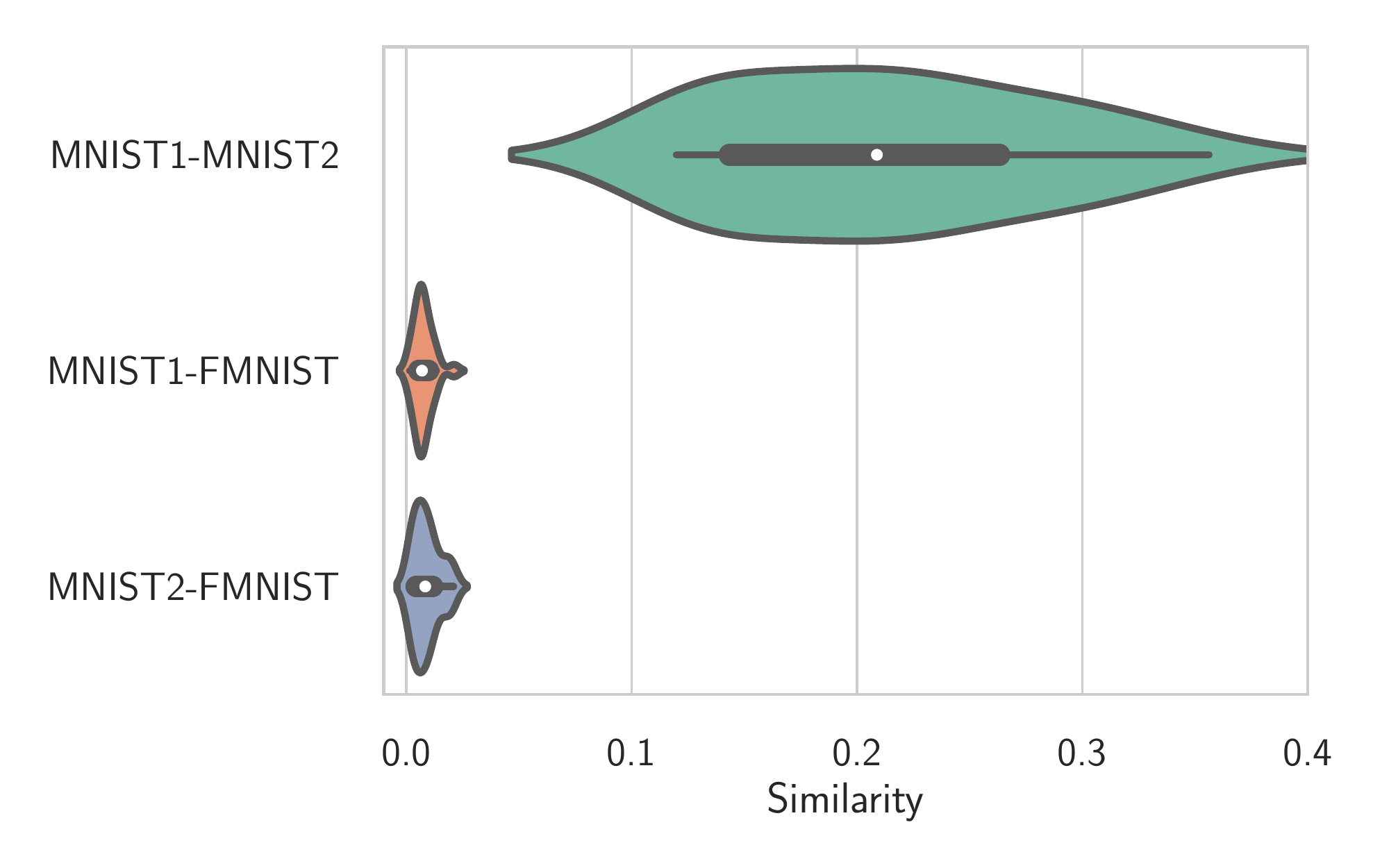}
	\caption{Cosine similarities of the Jacobians of trained LeNet-3s on the first half of MNIST (MNIST1). Models trained on FMNIST and evaluated on MNIST1 are less similar to models trained directly on MNIST1 than those trained on the second half of MNIST (MNIST2). The white dot in the middle is the average similarity across $25$ random seeds, with the black lines showing the quartiles. The width of the violin for each comparison corresponds to the density estimate across these seeds, showcasing a population.
	}
	\label{fig:jacobian_tasks}
\end{figure}

\begin{figure*}[t!]
	\centering
	\includegraphics[width=\textwidth]{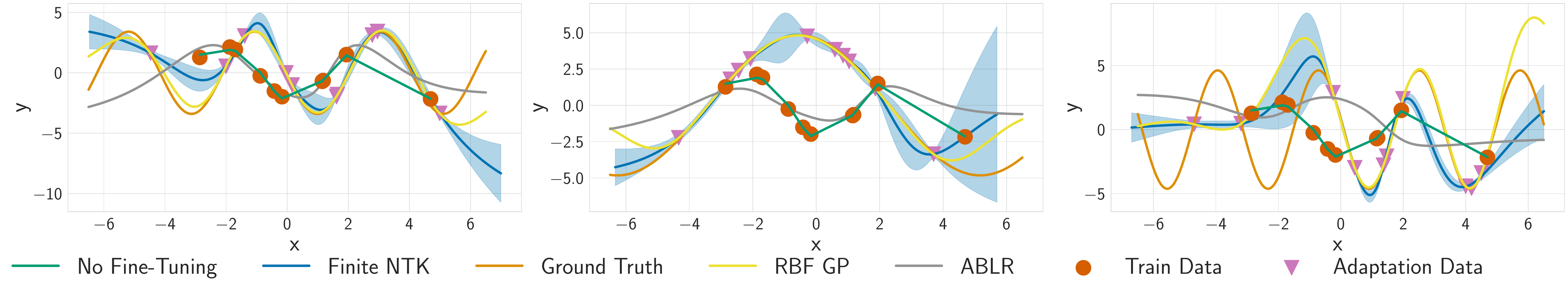}
	\caption{Posterior predictions on a few shot regression task using the finite NTK (blue). We performed training on the orange points, while adapting to the purple points (context), viewing each function as an independent draw from the GP described in Section \ref{sec:probmodel}. The comparisons here are {RBF GPs fit on the source task with the hyper-parameters shared across tasks (yellow)} and adaptive Bayesian linear regression (ABLR) (gray) \citep{perrone_scalable_2018}. Our finite NTK approach improves upon the RBF kernel, and significantly outperforms ABLR.}
	\label{fig:fewshotreg}
\end{figure*}

\paragraph{Transferability of Jacobian Features:}
Thus far, we have demonstrated that the Jacobian matrix is useful merely for a single task --- e.g. 
 a linearized model is a good inductive bias for both classification and regression.
 However, we next demonstrate that for a given architecture, the Jacobian matrix is similar across related tasks.
We consider trained networks on three different datasets --- two halves of MNIST (MNIST1 and MNIST2) \citep{lecun_gradient-based_1998}, and FMNIST \citep{xiao_fashion-mnist_2017}.
MNIST1 and MNIST2 are similar because they come from the same distribution, so we 
expect that classifiers trained on MNIST1 and MNIST2 will be similar due to being trained on similar data. 
By the same token, we expect these classifiers to have different representations than a classifier trained on FMNIST, which is images of clothing.

To test our hypothesis, we trained $25$ LeNet-3 \citep{lecun_gradient-based_1998} networks each on the three datasets, and then computed the full Jacobian matrix of each model on $5000$ images from MNIST1\footnote{We fine-tuned the final layer of the FMNIST networks to remove distinctions in class labels and predictions.}. Next, we computed the similarity of the matrices via their squared cosine similarity, e.g. $\text{sim}(A, B) = \frac{tr(A^\top B B^\top A)}{||AA^\top||_F ||BB^\top||_F}.$
We show histograms of the similarities across the $25$ comparisons in Figure \ref{fig:jacobian_tasks}, finding that the Jacobians are vastly more similar between classifiers trained on MNIST1 and MNIST2 (shown in blue), while the classifiers trained on MNIST1 and MNIST2 have nearly zero similarity to those from FMNIST (shown in orange and green).
This result supports our hypothesis, suggesting that the Jacobians of two models should be similar to each other if they were trained on similar enough tasks; full training details and the spectrum of the Jacobians for all of the models is shown in Appendix \ref{app:jac_similarity}.

\section{TRANSFERABILITY EXPERIMENTS}\label{sec:exps}
Finally, we demonstrate that, on a variety of regression and classification tasks, our linearization framework enables transfer of the inductive biases from one task to others and enables good representations of uncertainty. Experimental details are in Appendix \ref{app:regression_exp_details}, while further results are in Appendix \ref{app:transfer_exps}.

\paragraph*{Sinusoids:}
In Figure \ref{fig:fewshotreg}, we generate $3$ datasets 
and train a three-layer MLP with tanh activations to completion only on the first dataset (the orange points). 
We plot the fit from the finite NTK in blue (the mean and the confidence region), showing the validation (context) points in purple.
We replicate the generative process of \citet{kim_bayesian_2018}, with various periods for the 
sinusoids. 
Here, we compare to a trained adaptive basis linear regression (ABLR) model
 \citep{perrone_scalable_2018} is shown in grey and to transferred RBF Gaussian processes (yellow).
ABLR requires many gradient steps to converge on the adaptation task as it continually re-trains the features of the last layer.
Additionally, the linearized model performs comparably to transferred hyper-parameter RBF GPs which are close to the gold standard on this task,
because the functions are 
smooth.
However, the linearized model gives reasonable posterior predictive \emph{distributions} like the GP.
In Appendix \ref{app:fewshot}, we additionally compare to transferring the deep kernel learning latent space, and adaptive deep kernel learning \citep{tossou2019adaptive}, finding again that these alternative approaches underfit.

\begin{figure*}[!t]
\begin{subfigure}{0.45\linewidth}
\centering
\includegraphics[width=\linewidth,clip,clip,trim=0cm 2cm 0cm 2cm]{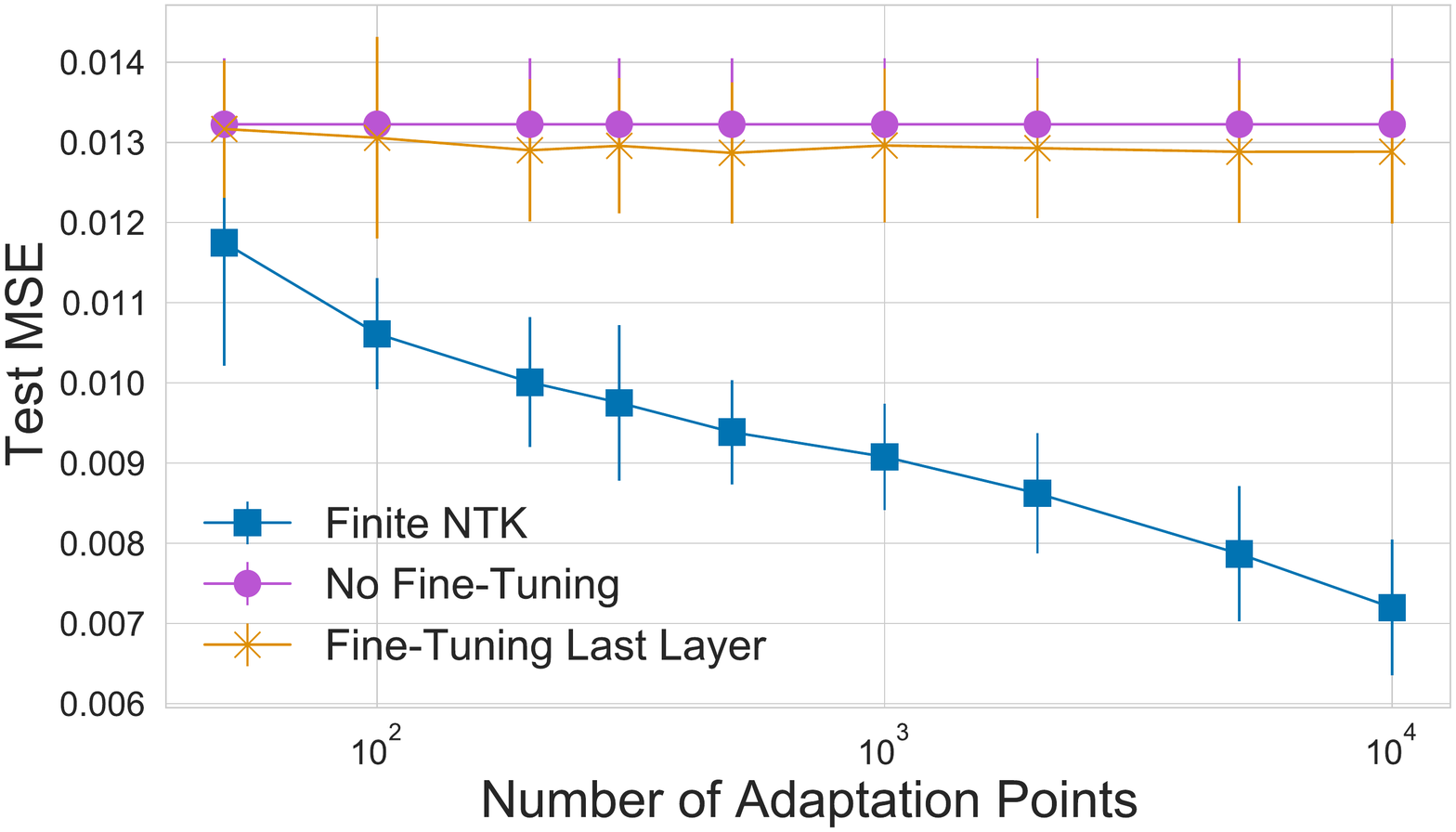}
\caption{Malaria dataset}
\label{fig:malaria}
\end{subfigure}
\begin{subfigure}{0.45\linewidth}
\centering
\includegraphics[width=\linewidth,clip,clip,trim=0cm 2cm 0cm 2cm]{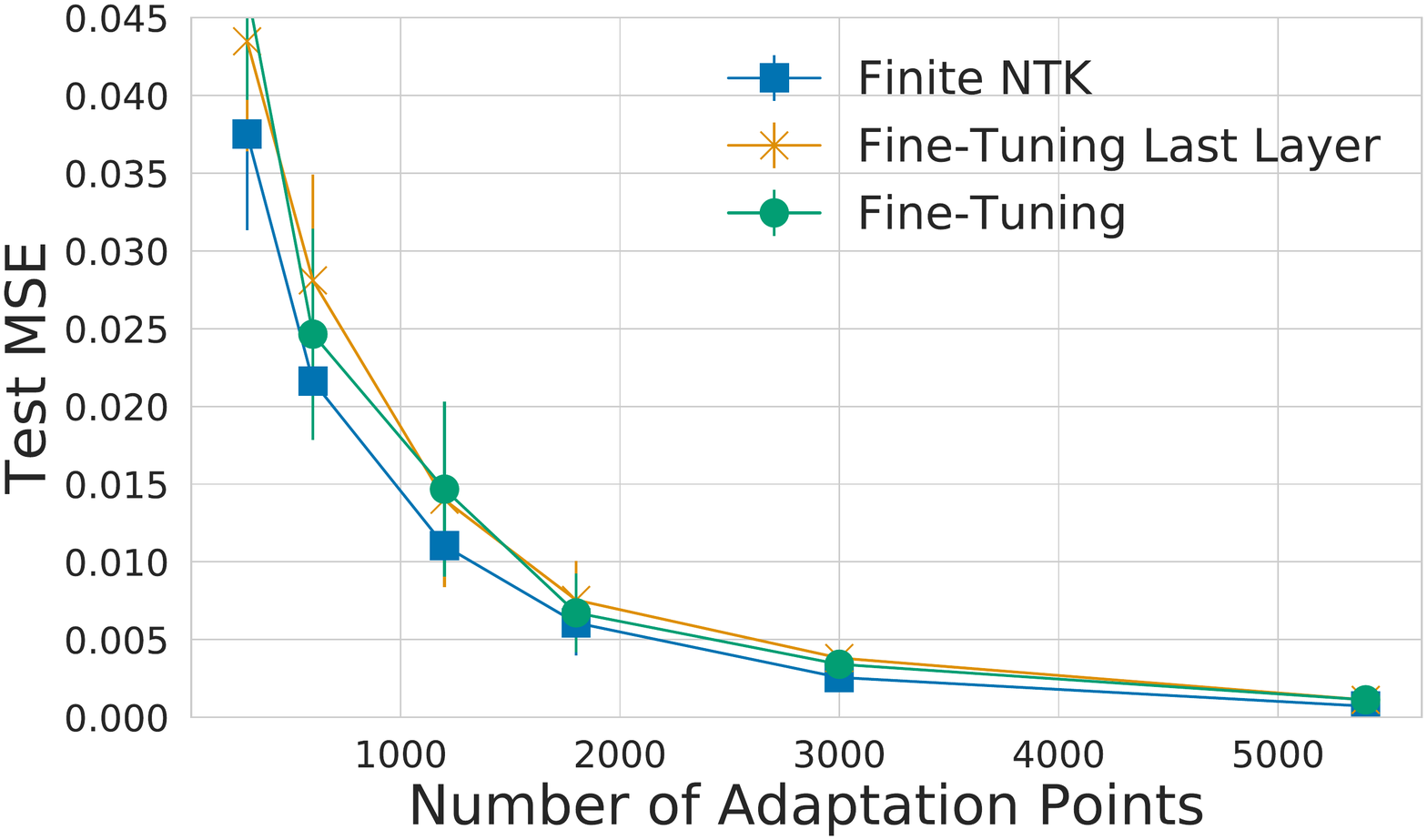}
\caption{Olivetti faces dataset}
\label{fig:olivetti}
\end{subfigure}
\caption{\textbf{(a)} Error on held out set for infection rate of \textit{Plasmodium falciparium} among African children. The NTK improves its accuracy on the held-out set as the number of validation data points from 2016 are given to it, while re-training the last layer stagnates and only has marginally better performance than no-retraining at all. \textbf{(b)} Error on held out set on the Olivetti fast adaptation task. In this setting, all methods improve with more adaptation points, and the NTK has the biggest advantage when there is a small amount of data for transfer.
}
\vspace{-0.2cm}
\end{figure*}
\begin{table*}[!h]
	\centering
	\caption{Test Accuracy on CIFAR10 for linearizing different layers of the BiGAN encoder. $*$ denotes numbers reproduced from Table 1 of \citet{mu_gradients_2019}, which are included as baselines. Using all of the layers produces slightly better results than using only the top layers. All linearization settings are best compared to the activation baseline of $62.87\%$ which consists of training a classifier on top of the feature extractor. Fine-tuning all of the layers is also better overall.}
	\label{tab:bigan_encoder}
	
	\begin{tabular}{|c | c | c | c | c | c |}
		\hline
		Layers & Top Layer & Top 2 Layers & Top 3 Layers & All Layers & All Layers (VI) \\
		\hline
		Fine-tuning& $71.78$* & $73.18$* & $74.30$* & $\emph{77.25}$ & - \\
		\hline
		Finite NTK & $70.28$*  & $71.08$* & $70.64$*  & $\emph{\textbf{72.65}}$ & $\emph{71.89}$\\
		\hline
	\end{tabular}
	\vspace{-0.4cm}
\end{table*}

\paragraph{Malaria Prediction:}
Inspired by \citet{balandat_botorch_2019}, we used data describing the infection rate of \textit{Plasmodium falciparium} (a malaria causing parasite) drawn from the Malaria Global Atlas\footnote{Extracted from {\scriptsize \url{https://map.ox.ac.uk}}.} in a domain adaptation task.
We trained a heteroscedastic neural network with three layers on 2000 data points from the 2012 map, before testing on varying numbers of data points from the 2016 map.
In Figure \ref{fig:malaria}, we show the test MSE for no re-training, the Jacobian as the kernel, and retraining the final layer versus 
the validation points given from the 2016 map.
The performance of the Jacobian kernel as a transfer model improves as the number of data points given to it 
at validation time 
increases.
For a fixed training budget (the same as the original training epochs), the re-trained last layer stagnates in performance.
By contrast, re-training the last layer for $6x$ the training epochs will give comparable results to the finite NTK, but at a significantly increased computational expense; see Appendix \ref{app:regression_exp_details} for further details.

\paragraph{Image Pose Prediction:}
We next convert the image rotation prediction task in \citet{wilson_deep_2015} on the publicly available Olivetti Faces dataset.\footnote{\scriptsize \url{scikit-learn.org/0.19/datasets/olivetti_faces.html}.} 
We used a similar version of the LeNet-3 architecture for this task \citep{lecun_gradient-based_1998}.
To construct this task, we used all 40 faces and rotated them such that $\theta \sim U(0, 30)$ where $\theta$ is the degree of rotation, cropping them to fit the $45 \times 45$ images, selected $20$ faces to serve as the training set and the other $20$ to serve as the adaptation set; the targets in all cases are the standardized degree of rotation. 
From the adaptation set, we randomly selected varying numbers of the data to serve as the adaptation points as we wish to have the methods quickly adapt to the new faces; we then evaluated on the MSE on the remainder of the adaptation points. 
We find, as displayed in Figure \ref{fig:olivetti}, that the linearized NN performs best across the entirety of the number of adaptation points, 
although the gains are small when many points are considered.

\paragraph{Transferring Unsupervised Models:}
We finally test the capability of the linearized model to work as a supervised model when the full model is trained on an un-supervised task, following the experiments of \citet{mu_gradients_2019}, who used BiGANs \citep{donahue2016adversarial} trained on CIFAR10 and linearized only the top convolutional layers.
We replicate their experimental setup, but consider inclusion of all of the features as well, by adding the Jacobian of the base feature extractor network into the model.
Like \citet{mu_gradients_2019}, we used a fixed projection from the output dimensionality of the network down to $10$ output dimensions so that we can perform classification with it.
The results are shown in Table \ref{tab:bigan_encoder}, where we see that in this problem the top layers seem to be at least as useful as the full gradient projections. Here, the all layers column refers to solely MAP trained models for direct comparison with \citet{mu_gradients_2019}, while all layers (VI) refers to our finite NTK model with variational inference.
We hypothesize that the underperformance of the finite NTK is due to using approximate inference.

\section{CONCLUSION}

We have shown how it is possible to build scalable kernel methods with the inductive biases of neural networks, but with closed-form training and no local optima, through a linearization. We combine these kernels with Gaussian processes for a principled representation of uncertainty. We show how to make inference scalable by developing efficient techniques for Jacobian vector products and conjugate gradients. The techniques we propose are in fact generally applicable, and in the future could be leveraged in a wide range of other settings where one wishes to perform efficient operations with a Jacobian or Fisher matrix. In this paper, we apply these ideas primarily for fast, convenient, and analytic transfer learning with neural networks. Exciting future work could further develop this direction for meta-learning, where one wishes to efficiently capture the transfer of knowledge across several tasks and uncertainty plays an important role.

\section*{Acknowledgements}
WJM, AGW are supported by an Amazon Research Award, NSF I-DISRE 193471, NIH R01 DA048764-01A1, NSF IIS-1910266, and NSF 1922658 NRT-HDR: FUTURE Foundations, Translation, and Responsibility for Data Science.
WJM was additionally supported by an NSF Graduate Research Fellowship under Grant No. DGE-1839302.
We would like to thank Jacob Gardner, Alex Wang, Marc Finzi, and Tim Rudner for helpful discussions, Jordan Massiah for help preparing the codebase, and the NYU Writing Center for writing guidance.

\bibliographystyle{apalike}
\bibliography{refs}

\appendix
\onecolumn
 \hsize\textwidth
\linewidth\hsize \toptitlebar {\centering
	{\Large\bfseries Supplementary Materials for \\ Fast Adaptation with Linearized Neural Networks \par}}
\bottomtitlebar 

\renewcommand\thefigure{A.\arabic{figure}}
\renewcommand\theequation{A.\arabic{equation}}   
\renewcommand{\thesection}{\Alph{section}}
\setcounter{equation}{0}
\setcounter{figure}{0}

\section{APPROXIMATE INFERENCE}\label{app:approxinf}
In this appendix, we discuss how we performed approximate inference in parameter space using the Jacobians as features.
In general, function space approaches either require computation of the diagonal of $\mJ_\theta^\top \mJ_\theta$ (for variational versions) or many more solves (for Laplace approximations) which would slow down inference considerably in our framework.

Using the finite NTK (or even the Taylor expansion perspective), we can replace the network, $f_\theta(\cdot)$ with the linearized model in the likelihood.
For example, for multi-class classification,
$p(y_i | \pmb{f}) = \text{Categorical}(y_i | \frac{\exp\{\pmb{f}_i\}}{\sum_{j=1}^C \exp\{\pmb{f}_j\}}),$
so that the linearized model is then given as
\begin{align}
	p_{\text{lin}}(y_i | x, \theta) = \text{Cat.}\left(y_i | \frac{\exp\{\mJ_\theta(x)^\top \theta'\}}{\sum_{i=1}^C \exp\{\mJ_\theta(x)^\top \theta'\}}\right).
	\label{eq:cat_lin}
\end{align}
For classification models, we will also consider the full Taylor expansion in Eq. \ref{eq:taylor} as a sanity check.

\paragraph{Stochastic Variational Inference (SVI)}
We tried stochastic variational inference in parameter space by using the ELBO and assuming a factorized Gaussian posterior.
Assuming the linearized loss in Equation \ref{eq:cat_lin}, the mini-batched ELBO \citep{hoffman_stochastic_2013} is simply
\begin{align*}
	\log{p(\vy | \mX)} &\geq \frac{N}{B} \sum_{i=1}^B \log{p(y_i | x_i, \theta)}- \text{KL}(q(\theta | \mu, \text{softplus}(v)) || \mathcal{N}(\theta | 0, \sigma^2 I)).
\end{align*}
We initialized $\mu = 0$ and $v = -5$ before training for ten more epochs.
If we include the predictions of the NN as well, we add in another mean term into the likelihood.

\paragraph{Laplace Approximation}
First, we can now utilize the Fisher information as a scalable Laplace approximation, so that (assuming a Gaussian prior with variance $\sigma^2$) the posterior is approximately $\mathcal{N}(\theta_i | \theta, (\mathbb{F}(\theta) + \sigma^2 I)^{-1}).$
Approximating the posterior in this manner will give rise to a degenerate version of the multi-class Laplace approximation derived in \citet[Chapter 3][]{rasmussen_gaussian_2008}.
To compute the approximation, we trained for 10 further epochs to get the MAP estimate.
At test time, we assumed conditional independence of the test points, computing the inverse of the Fisher information using the test batch via the approximate Fisher vector products that we derived previously upscaling this matrix by a term of $N/B.$

Under both linearizations, note that if $\theta$ is used as both the parameters for the Jacobian and the parameters in the model, then we can additionally write down the Fisher information matrix of this model and see that the only difference is in the function value of the inner portion of the loss function, which under the linearization assumption is assumed to be close to the true model, suggesting that we can use the fast Fisher vector products plus a Lanczos decomposition to sample from the Gaussian posterior.
We use a single sample at test time.

\section{SCALABLE EXACT GAUSSIAN PROCESSES}\label{app:fast_gps}

The chief bottleneck in Gaussian process computation is the cubic scaling as a number of data points due to having to invert the kernel matrix for posterior predictions, e.g. finding $K^{-1} z.$
However, \citet{gardner_gpytorch:_2018} provides a numerical linear algebra way around this issue via iterative methods; here, we give a more detailed explanation of their approach in the context of our use case.
To reduce the time complexity of solving linear sytems, we can use either the Lanczos algorithm or conjugate gradients \citep[Chapters 6-10]{saad_iterative_2003}; both of which use Krylov subspaces of a symmetric matrix $A \in \mathbb{R}^{p \times p}.$
The Krylov subspaces method takes as input a starting vector $b,$ before computing successive matrix vector products:
\begin{align*}
	\mathcal{K}(A, b) = \text{span} \left[ b, A b, A^2 b, A^3 b, \cdots A^{m} b \right].
\end{align*}
Orthogonalization and normalization (e.g. Gram-Schmidt) are then used to produce a basis matrix, $Q,$ while the resulting coefficients from the orthogonalization can be stored into a matrix $T_m,$ producing the recursion $A Q_{m} = Q_{m+1} T_m.$ As $A$ is symmetric, $T_m$ is tri-diagonal, producing the decomposition: $A \approx Q_m T_m Q_m^T.$
Taking $m = p,$ produces the decomposition, $A = Q T Q^T.$
Note that solves can be produced from Lanczos as $A^{-1} b \approx ||b||Q_m T_m^{-1} e_1,$ while the conjugate gradients method directly returns a closely related solution.
Finally, we can see that only a single (or possibly a constant number) of matrix vector multiplies with $A,$ reducing the complexity to $\mathcal{O}(p^2 m).$
Conjugate gradients converges exactly when $r = p,$ but that we can speed up the convergence rate by using a pre-conditioner, $P,$ where $P \approx A^{-1}.$ We use the standard pivoted Cholesky preconditioner \citep{gardner_gpytorch:_2018}.

Finally, for predictive variance computations, we must store  $(\mJ_\theta \mJ_\theta^T + \sigma^2 I)^{-1}$ or $(\mJ_\theta^T \mJ_\theta + \sigma^2 I)^{-1}.$ However, we can approximate this by an approximate eigen-decomposition using the Lanczos decomposition \citep[e.g.][]{pleiss_constant-time_2018}, writing $(\mJ_\theta \mJ_\theta^T + \sigma^2 I)^{-1} \approx RR^T,$ where $R = QT^{-1/2}.$
This requires only $\mathcal{O}(pr)$ storage and again can be computed in quadratic time as $T$ is tri-diagonal.

\section{FURTHER DISCUSSION OF THE FISHER INFORMATION MATRIX}\label{app:fast_fvp}

\subsection{Relating the Fisher Information Matrix to the NTK}

\paragraph{Regression:} An interesting example is given by homoscedastic regression with a Gaussian likelihood, where $y \sim \mathcal{N}(f(x), \sigma^2 \mI).$
There, the empirical Fisher information matrix is $\frac{1}{n} \mJ_\theta \mJ_\theta^T,$ while the neural tangent kernel (and the linearization) is $\mJ_\theta^T \mJ_\theta.$
The Fisher information and the finite NTK as we use it then have the same eigenvalues (up to a constant factor of $n$) as they are similar matrices. Similar connections between the Jacobian and the Fisher information matrix are used by \citet{tosi_metrics_2014}, and the connection seems to originate in the generalized Gauss-Newton decomposition of \citet{golub_differentiation_1973}. \citet{yang_scaling_2019} first noted the connection in the context of the neural tangent kernel at infinite width.

\paragraph{General Losses:}
For general losses, it is still possible to relate the Fisher information matrix to either the Gram matrix or the NTK.
\begin{align*}
	\mathbb{F}(\theta) :&= \mathbb{E} (\nabla_\theta \log p(y | x, \theta) \nabla_\theta \log p(y | x, \theta)^\top )\\
	&= \mathbb{E}(\nabla_\theta f(x; \theta) \nabla_f \log p(y | f) \nabla_f \log p(y| f)^\top \nabla_\theta f(x; \theta)^\top) \\
	&= \mathbb{E}_{p(x)}(\nabla_\theta f(x; \theta) \mathbb{E}_{p(y|f)} (\nabla_f \log p(y | f) \nabla_f \log p(y| f)^\top) \nabla_\theta f(x; \theta)^\top) \\
	&\approx \mJ_\theta \mH_\theta \mJ_\theta,
\end{align*}
with the approximation coming as the Jacobian is computed over the observed dataset rather than the true data distribution --- exact if we use the empirical Fisher information, taking the empirical data distribution to be the tue data distribution, $p(x).$ $\mH_\theta$ is the matrix of the gradient covariance with respect to the inputted function $\mH_\theta := \mathbb{E}_{p(y|f)} (\nabla_f \log p(y | f) \nabla_f \log p(y| f)^\top).$
Note that $\mH_\theta(x)$ is block-diagonal and positive semi-definite if the likelihood can be written to factorize across data points (e.g. the responses are i.i.d).
We can parameterize the empirical Fisher in terms of the eigen-decomposition, $\mathbb{F}(\theta) \approx \mJ_\theta \mH_\theta \mJ_\theta^T = S\Lambda S^T.$
Future work is necessary to be able to efficiently exploit the decomposition beyond the simple parameter space Laplace approximation we used.

\subsection{Fast Fisher Vector Products via Directional Derivatives}
To implement the fast Fisher vector products as described in Section \ref{sec:computational}, we need to know the closed form of the KL divergence of the likelihood distribution to itself (e.g. the KL between two Gaussians for regression).
For many probability distributions used in machine learning, the KL divergence is closed form and often pre-implemented (for PyTorch in \texttt{torch.distributions}).
For fixed homoscedastic Gaussian noise, the KL divergence is
\begin{align*}
	\KL(p(y\mid\theta)||p(y\mid\theta')) = \frac{(f_\theta - f_{\theta'})^2}{2 \sigma^2}.
\end{align*}
\begin{wrapfigure}{l}{0.5\textwidth}
	\centering
	\includegraphics[width=0.5\textwidth]{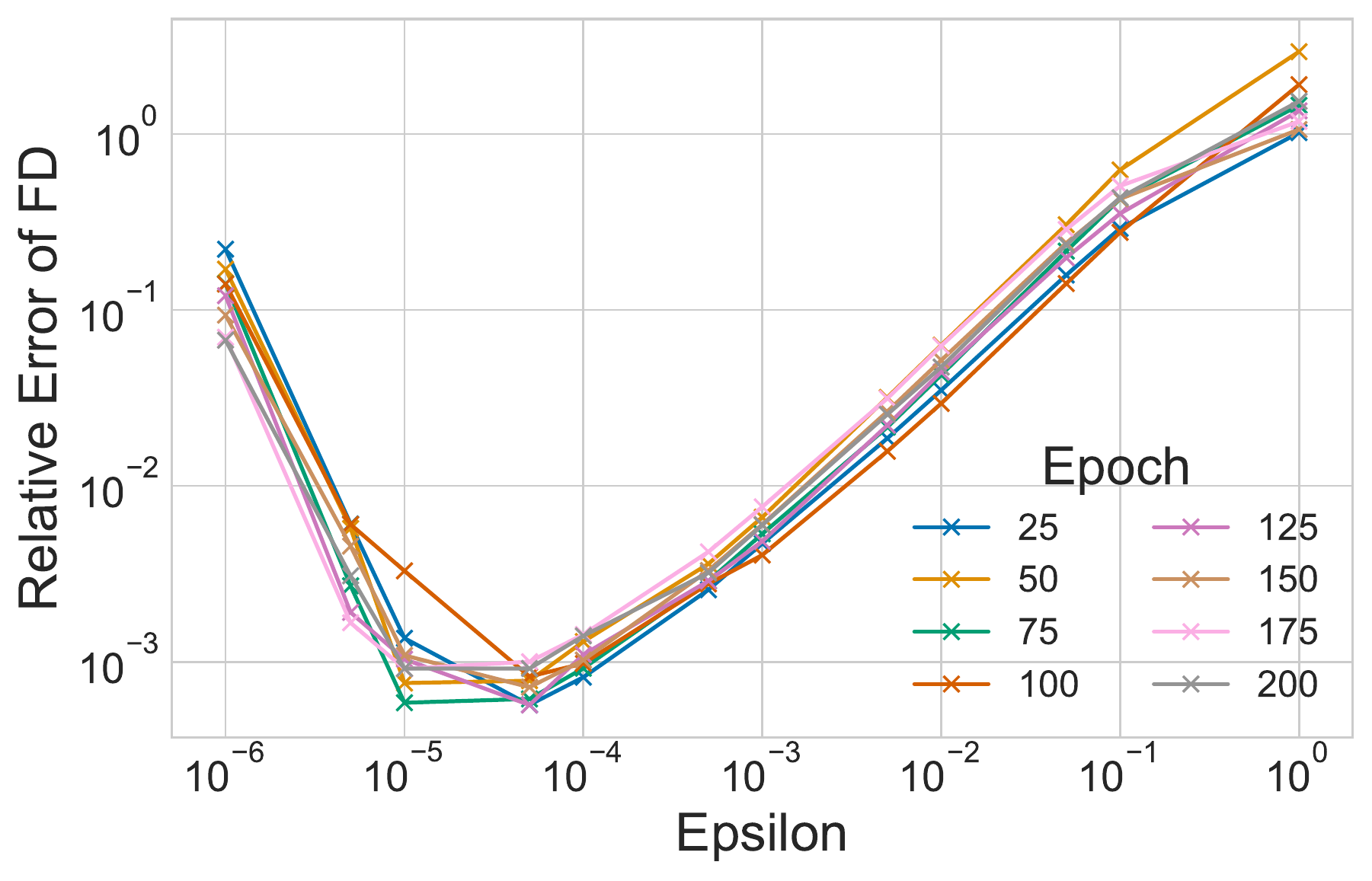}
	\caption{Accuracy of Fisher-vector products as a function of tuning parameter $\epsilon$ for finite differences (FD, our approximate fast Fisher-vector product) versus autograd (AG), which is exact, across training of a PreResNet56 measured every $25$ epochs. The finite differences approximation is accurate to a relative error of less than 1\% for most choices of $\epsilon.$}
	\label{fig:fvp_accuracy}
	\vspace{-0.5cm}
\end{wrapfigure}
For multi-class classification with categorical likelihoods, the KL divergence reduces to a summation over all classes, giving
\begin{align*}
	\KL(&p(y\mid\theta)||p(y\mid\theta')) \\
	&= \sum_{c=1}^C \text{detach}(p(y_c\mid\theta)) \log\left(\frac{p(y_c\mid\theta)}{p(y_c\mid\theta')}\right),
\end{align*}
where $\text{detach}(.),$ refers to an operation that will not play a role in the computation graph.

Finally, we show the approximation error between the directional derivative as defined in Eq. \ref{eq:kl_deriv_fvp} and the exact $F_iv$ computed using the standard second order autograd; this is shown in Figure \ref{fig:fvp_accuracy}.
We used a modern neural network architecture, PreResNet56, on the benchmark CIFAR10 dataset and computed the relative error as a function of $\epsilon$:
\begin{equation*}
	\text{Err}(\epsilon):=\frac{||(F_i \nabla f(x))_{AG} - (F_i \nabla f(x))_{FD(\epsilon)}||}{||(F_i \nabla f(x))_{AG}||},
\end{equation*}
through various stages of the standard training prodcedure with stochastic gradient descent.
Crucially, we note that the relative error produced by this approximation is on the order of $1e-3$ and stays nearly constant throughout training, suggesting a simple procedure for tuning this hyper-parameter at the beginning of training.
Furthermore the approximations fail gracefully, only decaying to large error when $\epsilon$ is too small due to numerical precision issues or when $\epsilon$ is too large and the finite diferences approximation is not accurate. For regression, we observed similar qualitative results with higher stability.

\section{FAST ADAPTATION MODEL}\label{app:fast_adaptation}

We define a deep neural network, $f,$ as taking an input, $x \in \mathbb{R}^d,$ and mapping it to an output, $y \in \mathbb{R}^o,$ with parameters $\theta \in \mathbb{R}^p,$ letting $\mathcal{D} = \{(x_i, y_i)\}_{i=1}^n.$ We will additionally describe the full dataset as $\mX = \{x_i\}_{i=1}^n$ and $\vy = \{y_i\}_{i=1}^n$.
For the purposes of fast adaptation, we consider an \emph{initial} task $t=0$ used to obtain parameters $\theta \triangleq \theta_{MLE}$ which are then re-used for learning the mappings $f_t$ for the rest of the tasks $t=1,\dots, T$. In this way, we rapidly adapt $f_0$ to $f_t$.
That is, we pre-train a network on the first task given the first set of data $(\mX_0, \vy_0)$ using any optimization procedure (e.g. SGD or Adam) to get out a set of parameters $\theta_{MLE}.$

In the multi-task learning scenario, we have multiple related tasks $t=1, \dots, T$, where a task is defined as learning a neural network $f_t$, which depends on parameters $\theta_t$, given the corresponding dataset $\mathcal{D}_t = \{ \mX_t, \vy_t \}$.
So for every subsequent task, we lazily compute the Jacobian for task t using the new training inputs (or context points), $\vx_t$ and then compute the predictive mean cache of the Gaussian process (e.g. the terms dependent on the training data in Equations \ref{eq:parameter_space_inference} and \ref{eq:gp_pred_dist}). 
For function space inference (Equation \ref{eq:gp_pred_dist}), following \citet{gardner_gpytorch:_2018} and \citet{pleiss_constant-time_2018}, we compute $m = \mJ_\theta(\mJ_\theta^\top \mJ_\theta + \sigma^2 \mI_n)^{-1} \vy$ and $RR^\top \approx \mJ_\theta (\mJ_\theta^\top\mJ_\theta + \sigma^2 \mI_n)^{-1}\mJ_\theta.$
On the task's test set, we then only have to compute $\mu(x^*) =	\mJ_\theta^{*T} m$ (ignoring the GP mean function) and $\sigma^2(x^*) = 	\mJ_\theta^{*T}	\mJ_\theta^* - \mJ_\theta^{*T} R R^\top \mJ_\theta^*.$
Parameter space inference uses the same mechanism, pre-computing $m =  (\mJ_\theta \mJ_\theta^\top + \sigma^2 \mI_p)^{-1}  \mJ_\theta \vy$ and $BB^\top \approx  (\mJ_\theta \mJ_\theta^\top + \sigma^2 \mI_p)^{-1},$
but the predictive variance becomes $\sigma^2(x^*) =\sigma^2 \mJ_\theta^{*\top}  BB^\top \mJ_\theta^{*}.$
In both cases, only a single Jacobian vector product on the test inputs $\vx^*$ is required to get a test predictive mean and variance.

\section{EXPERIMENTAL DETAILS}\label{app:regression_exp_details}

\subsection{Similarity of Jacobian across Tasks}\label{app:jac_similarity}
We used the LeNet implementation from \url{https://github.com/activatedgeek/LeNet-5} and followed their training procedure, resizing the images to be $32 \times 32,$ training for $20$ epochs with a learning rate of $2e-3$ using Adam and a batch size of $256.$.
For MNIST1 and MNIST2, we fixed the seed and then split the dataset using the first $30000$ images.
For FashionMNIST, we re-trained the final linear layer (so as to not disturb the internal representations) for $1000$ steps using Adam with a learning rate of $0.01.$

\begin{wrapfigure}{l}{0.5\textwidth}
	\centering
	\includegraphics[width=0.5\textwidth]{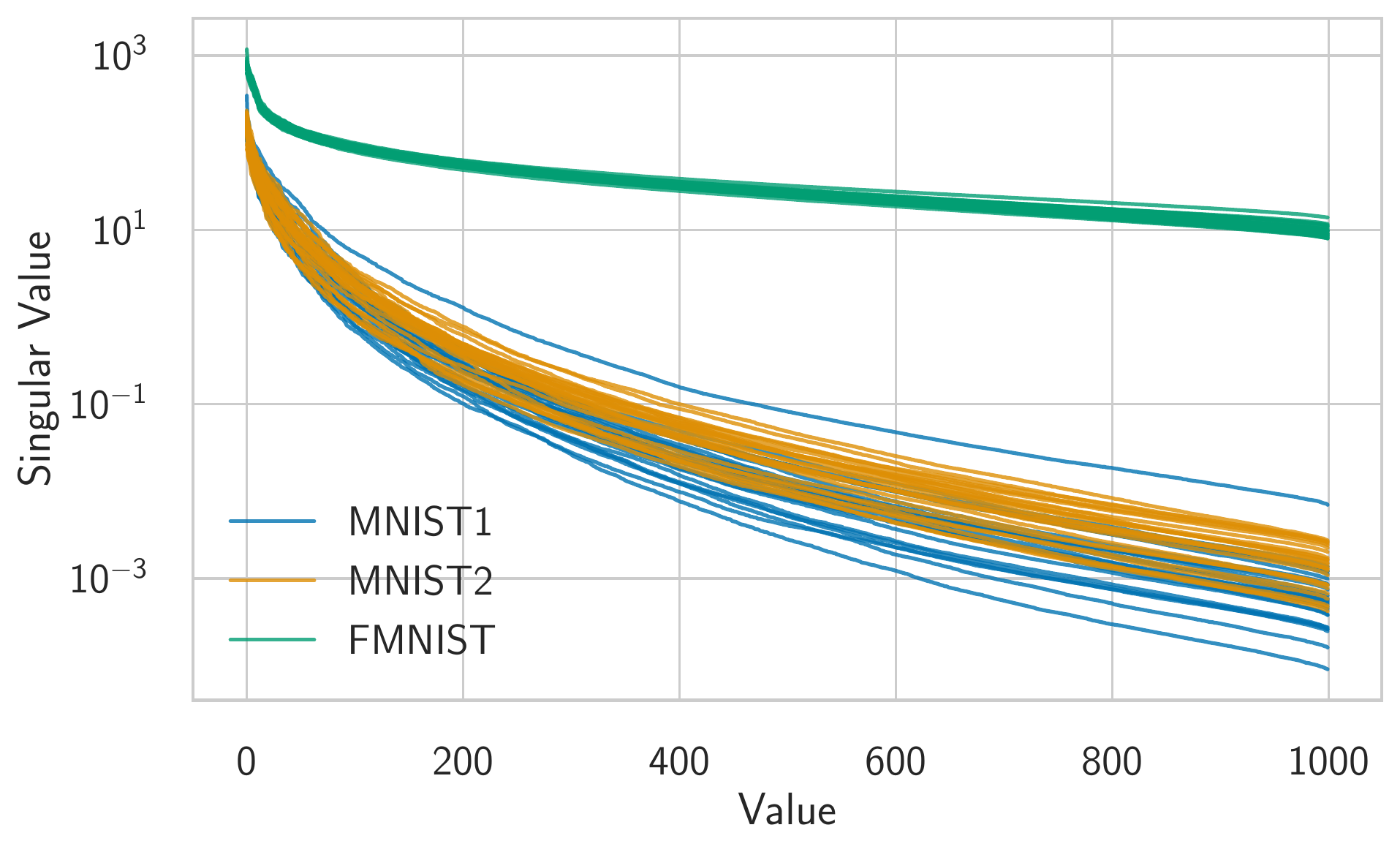}
	\caption{Singular values of the Jacobian on $5000$ images from the first half of MNIST (MNIST1) for$25$ models trained each on MNIST1, as well the second half of MNIST (MNIST2) and on Fashion MNIST (FMNIST). The singular values decay much more rapidly (and in a similar fashion) for models trained on MNIST1 as well as MNIST2, implying that their corresponding finite NTK kernels will have similar properties. }
	\label{fig:jacobian_tasks_v2}
	\vspace{-1cm}
\end{wrapfigure}

In Figure \ref{fig:jacobian_tasks_v2}, we show $1000$ singular values\footnote{Using \texttt{torch.svd\_lowrank}.} of the Jacobian for data coming from the MNIST1 task (first half of MNIST) from $25$ models each that were pre-trained on MNIST1, the second half of MNIST (MNIST2), or Fashion MNIST. 
To ensure fair comparison across tasks, we re-trained the classifier networ of Fashion MNIST until it had small training error on MNIST1. 
Surprisingly, the Jacobians of the models trained on MNIST2 have similar singular values to the Jacobians of models trained on MNIST1, despite being trained on different data. 

As our finite NTK kernel is given by $\mJ_\theta \mJ_\theta^\top,$ the squared singular values of $\mJ_\theta$ are the eigenvalues of the kernel.
The eigenvalues of kernel matrices help to determine the properties and inductive biases of the kernel, so that we expect two kernels with similar eigenvalues (and a similar decay rate for them) to have similar properties.

\subsection{Regression and Classification Details}
For all regression experiments, unless otherwise specified, we set $\sigma^2$ to be the training mean squared error on the original source task. We could alternatively have marginalized or optimized it.

\paragraph{Qualitative Regression Experiments}
For \ref{fig:increasing_width}, the true regression function is $y = 0.1 x^2 + |x| + \epsilon_i, $ with $\epsilon_i \sim \mathcal{N}(0, 0.9^2),$ and the 300 training data points are drawn from $\mathcal{N}(\pm 5, 0.8^2).$ The neural networks were trained with 250 epochs of SGD with momentum with learning rate $1e-4$ and momentum $0.9.$ 
For Figure \ref{fig:different_architectures}, we next fit three different architectures with tanh activations --- $[2110, 3, 2110]$ ($21104$ parameters), $[10, 1000, 10]$ ($21041$ parameters) and $[100, 100, 100]$ ($20501$ parameters) --- on $150$ data points with the same x generation process as before, but now $y = \sin(3x) + 1.5 \sin(0.5 x) + 0.8 |x| - 4 + 0.1 * \mathcal{N}(0, 0.1^2).$

\paragraph{Sinusoidal Regression}
The generative process matches the generative process of \cite{kim_bayesian_2018}, where we generate $x \sim \mathcal{N}(0, \mI_{10})$ and then $y_i = A \sin(w x_i + b) + \epsilon_i,$ where $A \sim U(0.1, 5.),$ $w \sim U(0, 2\pi)$ and $\epsilon \sim \mathcal{N}(0, 0.01A).$
We follow the setup of \cite{kim_bayesian_2018}, and use neural networks with two hidden layers and 40 hidden units and tanh activations (or ReLU for Figure \ref{fig:fewshotreg_relu}).
On the first task, we train the network with batch sizes of 3 for 2500 epochs using stochastic gradient descent with a learning rate of 1e-3 and momentum = 0.9.
We then incorporate this into a NTK model and compute the predictive variances using either function or parameter space.
For Figure \ref{fig:fewshotreg_fisher}, we perform parameter space inference with our novel implementation of Fisher vector products ($\epsilon = 1e-4.$). 

\paragraph{Malaria}
For the Malaria global atlas experiment, we trained a single neural network with three hidden layers ($2 - 500 - 500 -2$) and tanh activations on $2000$ randomly selected datapoints of the 2012 map in Nigeria (using the dataset preprocessing from \citet{balandat_botorch_2019}), training with a heteroscedastic loss, so that the likelihood model was $\mathcal{N}(y | \mu_\theta(x), 1e-5 + \text{softplus}(\sigma_\theta(x))).$ We trained using SGD with momentum with batch sizes of 200 for 500 epochs decaying the initial learning rate from 1e-3 by a factor of 10 every 100 epochs.
For the fine-tuned final layer, we continued training for 100 more epochs using the same procedure; we note that slightly better performance was acheived by training for 1000 epochs; however, this rapidly becomes an unfair comparison due to the increased training time. After all, if training for 1000 epochs, why not just train a full model?
We used inference in function space here as it was slightly more numerically stable (e.g. a smaller conjugate gradients residual norm) for small amounts of data.

\paragraph{Olivetti}
For the olivetti faces dataset, we trained neural networks similar to the LeNet-3 architecture used in \citep{wilson_deep_2015}; that is, after enforcing that the image was $45 \times 45,$ a convolutional layer with kernel $5 \times 5$ and $20$ channels, another with the same kernel size and $50$ channels, then a max pooling layer, and linear layers of $3200-500-2$ with ReLU activations. We again trained with the same heteroscedastic loss, but here used Adam with a learning rate of $1e-3,$ a batch size of $32,$ and for $150$ epochs. 
In this setting, we found that re-training the \emph{entire} neural network seemed practical due to the small size of the dataset, so we included it in the experiments.
Again, we used inference in function space here as it was slightly more numerically stable, except for the largest data point size, where we used inference in parameter space for numerical stability.

\paragraph{Linearized PreResNets}
To train the PreResNets, we followed the training procedure and model definitions originally from \url{https://github.com/kuangliu/pytorch-cifar}, except that we varied depth, used random cropping and horizontal flips, training with SGD with momentum for $200$ epochs with an initial learning rate of $0.1$ and weight decay of $5e-4.$
For the linearization experiments, we compared to accuracies at the end of training, training the approximate inference models with an initial learning rate of $1e-3$ and using Adam for $10$ epochs. The prior was again set to $\mathcal{N}(0,1);$ here, we found that the prior variance was important, as high variances performed poorly.
We followed the same procedure for transferring the models to STL10, described in Appendix \ref{app:transfer}.
For fine-tuning those networks, we used the same learning rate, optimizer, and training time.

\section{FURTHER EXPERIMENTS}\label{app:transfer_exps}
\subsection{PreResNets: Linearization on CIFAR10}
\begin{figure}[!htb]
	\centering
	\includegraphics[width=\linewidth]{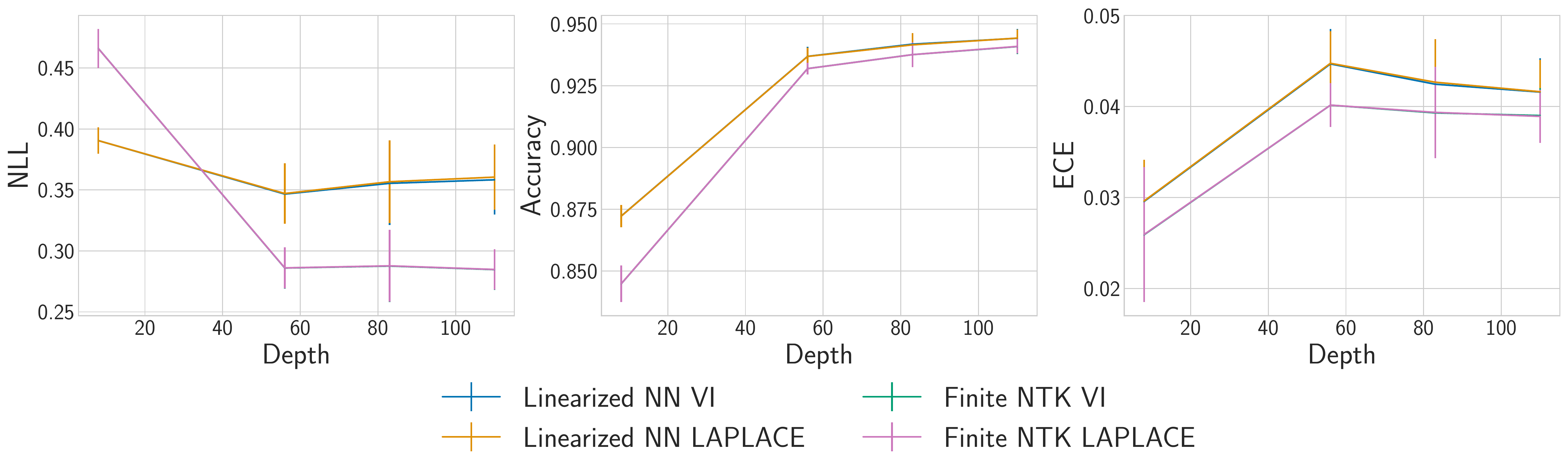}
	\caption{Test Accuracy of PreResNets averaged over $10$ random seeds. Both VI and Laplace are included here to see if the approximate inference method mattered much for either the linearized NN or the finite NTK. They perform similarly with some small differences.}
	\label{fig:prn_c10_approxinf}
\end{figure}
In Figure \ref{fig:prn_c10_approxinf}, we show the differences between using variational inference and the Laplace approximation for linearized NNs and the finite NTK on NLL, accuracy, and the expected classifcation error (ECE). 
Somewhwat surprisingly, we found that there was not that much difference; however, we attribute this to using the mean parameter for SVI at test time, and the fact that the Laplace approximation produces essentially the same mean parameter due to training with SGD before using a single sample at test time.
MAP training produced similar results; except that it was faster than the Laplace approximation at test time.

\subsection{PreResNets: CIFAR-10 to STL10}\label{app:transfer}
\begin{figure}[!htb]
	\centering
	\includegraphics[width=\linewidth]{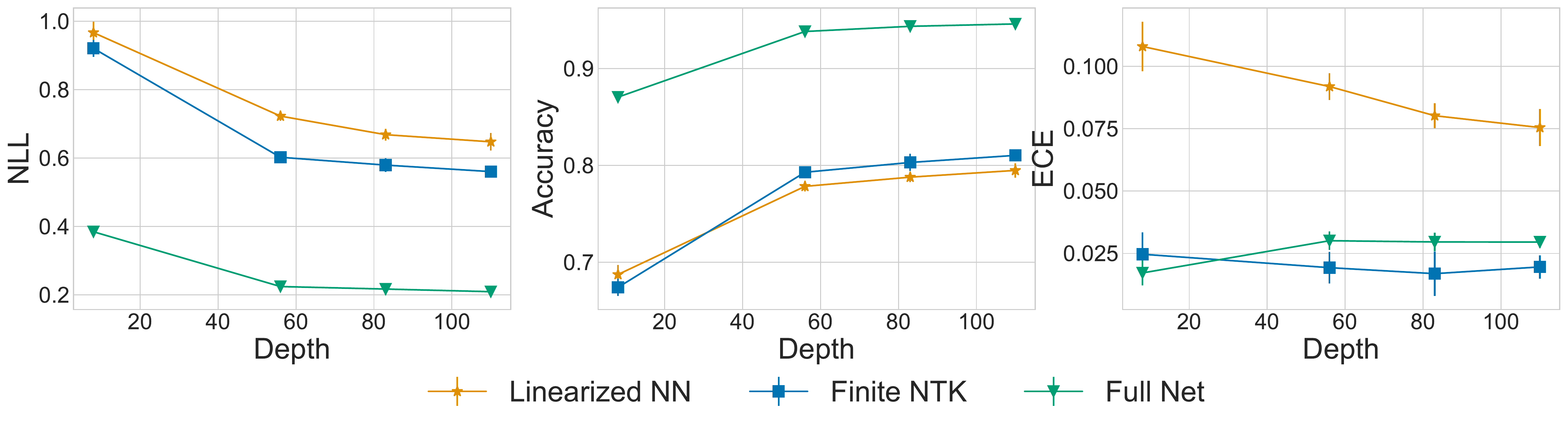}
	\caption{Test Accuracy of PreResNets averaged over $10$ random seeds with fine-tuning (green), the linearized NN (blue) and the finite NTK (orange) after training on CIFAR-10 and transferring to STL10. Linearization using solely the Jacobian as the features (yellow) is competitive with fine-tuning the final layer and no fine-tuning.}
	\label{fig:prn_c10_to_stl10}
\end{figure}

Finally, we perform domain adaptation by training PreResNets on CIFAR-10 \cite{krizhevsky_learning_2009}, and transfer classification to STL10 \cite{netzer_reading_2011}. To do so, we downsample STL10 to utilize 32 x 32 images.\footnote{Nine of the ten CIFAR10 classes are represented in STL10. We map these classes to each other and the non-matching ones as well.} 
We show accuracy as a function of depth for linearized preactivation ResNets, comparing to fine-tuning and no fine-tuning, in Figure \ref{fig:prn_c10_to_stl10}.
Here, approximate inference with Laplace approximations converged but again the VI results for the linearization did not.
Interestingly, the deeper Laplace approximations that incorporated the function itself had higher variance, possibly due to over-fitting, while the Laplace approximation using the Jacobian as the features nearly matched the performance of the non-fine tuned model on this task.
Specifically, we find that utilizing the full Taylor expansion and approximate inference is somewhat better than no adaptation, while using the Laplace approximation without the prediction of the model is a competitive baseline for a linear model.

\subsection{Further Sinusoids Plots}\label{app:fewshot}

\begin{figure*}[h]
	\centering
	\includegraphics[width=\textwidth]{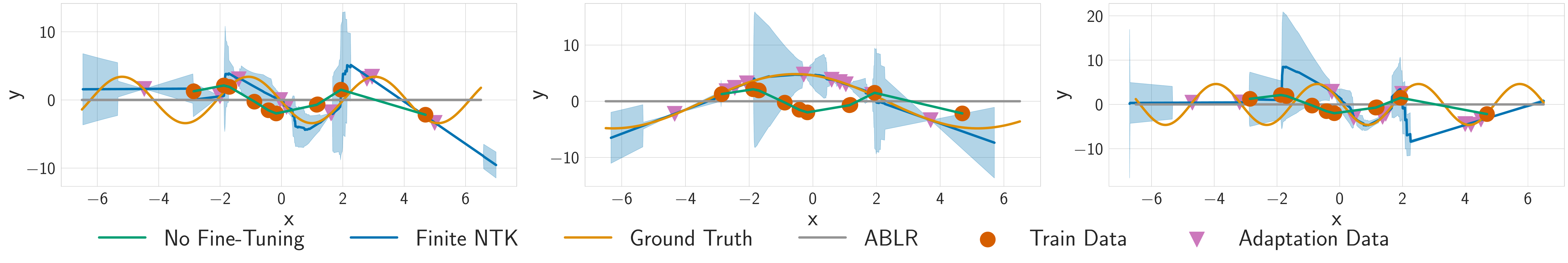}
	\caption{Posterior predictions on a few shot regression task produced with the NTK as the kernel using a network with ReLU activations. Here, the ReLU network is a poor inductive bias for this task, as is reflected in the jagged predictive variances and in the ABLR comparison which is completely flat.}
	\label{fig:fewshotreg_relu}
\end{figure*}

\begin{figure*}[h]
	\centering
	\includegraphics[width=\textwidth]{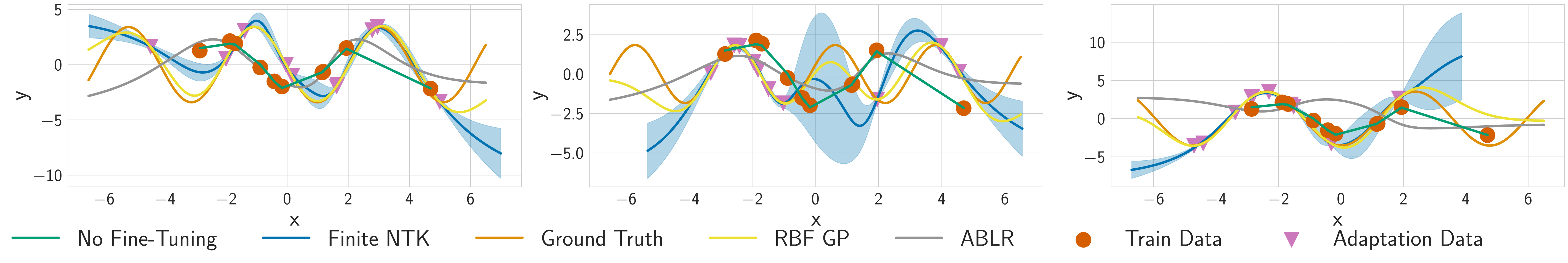}
	\caption{Posterior predictions on a few shot regression task produced with the NTK as the kernel using Fisher vector products instead of Jacobian vector products. These results are nearly identical to the results with Jacobian vector products described in the main text.}
	\label{fig:fewshotreg_fisher}
\end{figure*}

In Figure \ref{fig:fewshotreg_relu}, we show transferring with ReLU activations and the accompanying sharp variance predictions (with the same width as described in Appendix \ref{app:regression_exp_details}).
In Figure \ref{fig:fewshotreg_fisher}, we show the same networks as in the main text but with the inference performed using Fisher vector products (e.g. in parameter space) instead of using Jacobian vector products. The results are essentially identical to those displayed in the main text, as expected.

\begin{figure*}[h]
	\vspace{-0.25cm}
	\includegraphics[width=\linewidth]{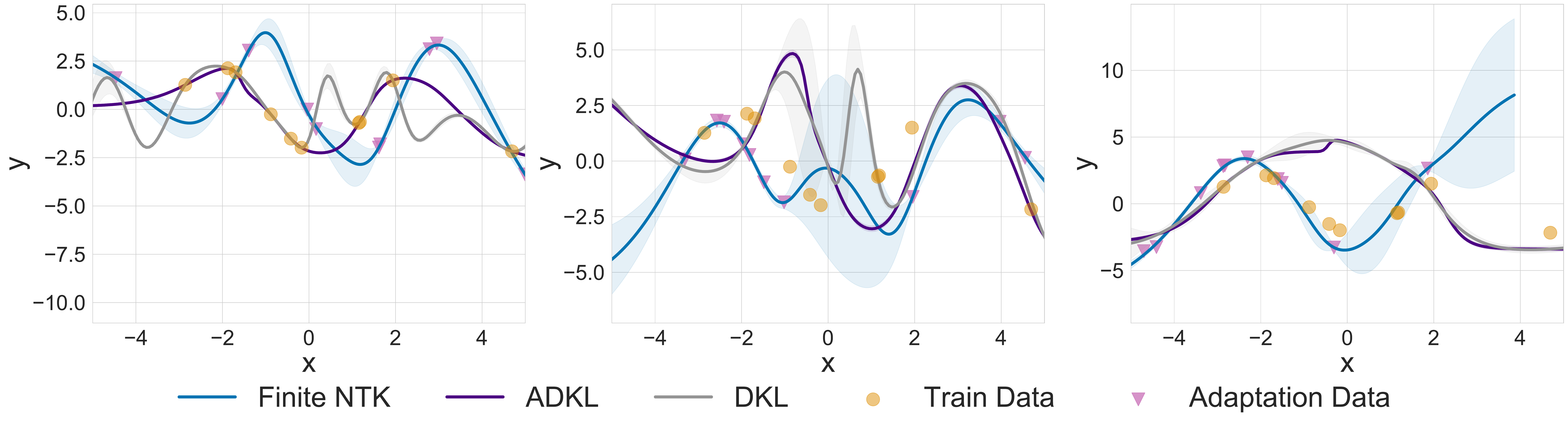}
	\caption{Comparison to ADKL \citep{tossou2019adaptive} and transferred DKL, a two task version of \citet{patacchiola2019deep} on the sinusoids problem. Both ADKL and transferred DKL underfit.}
	\label{fig:fig_adkl}
	\vspace{-0.4cm}
\end{figure*}

Finally, in Figure \ref{fig:fig_adkl}, we compare to adaptive deep kernel learning (ADKL) \citep{tossou2019adaptive} and a deep kernel learning implementation that transfers the learned representation into a new Gaussian process model, with the same hyper parameters.
The deep kernel learning implementation is a two task version of \citet{patacchiola2019deep}.
Both ADKL and the transferred DKL significantly underfit on the second task for sinusoids, which indicates their unsuitability for few shot transfer.
Both methods are designed for meta-learning over many tasks instead.

\end{document}